\documentclass[sigplan,screen]{acmart}
\setcopyright{acmlicensed}
\acmPrice{15.00}
\acmDOI{10.1145/3453483.3454083}
\acmYear{2021}
\copyrightyear{2021}
\acmSubmissionID{pldi21main-p375-p}
\acmISBN{978-1-4503-8391-2/21/06}
\acmConference[PLDI '21]{Proceedings of the 42nd ACM SIGPLAN International Conference on Programming Language Design and Implementation}{June 20--25, 2021}{Virtual, Canada}
\acmBooktitle{Proceedings of the 42nd ACM SIGPLAN International Conference on Programming Language Design and Implementation (PLDI '21), June 20--25, 2021, Virtual, Canada}
\startPage{1}

\usepackage{booktabs}   
\usepackage[skip=5pt]{caption}
\usepackage{subcaption} 

\usepackage{flushend}
\newcommand{\projectname}{DNNFusion\xspace}
\usepackage[noend,ruled,vlined]{algorithm2e}
\usepackage{multicol}
\usepackage{multirow}
\usepackage{makecell}
\usepackage{array}
\usepackage{amsmath}
\usepackage{ctable}
\usepackage{enumitem}
\usepackage{pifont}
\newcolumntype{P}[1]{>{\centering\arraybackslash}p{#1}}

\definecolor{wniu}{RGB}{255, 103, 104}

\definecolor{gagrawal}{RGB}{148, 33, 147} 

\newcommand{\bren}[1]{\textcolor{red}{#1}}
\definecolor{bren}{RGB}{255, 38, 0}

\newcommand{\compactparagraph}[1]{{\bf {\em {\noindent #1}}}}

\usepackage{listings}
\usepackage{xcolor}
\definecolor{codegray}{RGB}{0,92,240}
\definecolor{codepurple}{rgb}{0.58,0,0.82}
\definecolor{keyword}{RGB}{186, 45, 162}
\lstdefinestyle{mystyle}{
    backgroundcolor=\color{white},   
    commentstyle=\color{codegray},
    keywordstyle=\bfseries\color{keyword},
    numberstyle=\tiny\color{codegray},
    stringstyle=\color{codepurple},
    basicstyle=\ttfamily\footnotesize,
    breakatwhitespace=false,         
    breaklines=true,                 
    captionpos=b,                    
    keepspaces=true,                 
    numbers=left,                    
    numbersep=5pt,                  
    showspaces=false,                
    showstringspaces=false,
    showtabs=false,                  
    tabsize=2,
    morekeywords={entry},
}
\RequirePackage[normalem]{ulem}
\newcommand{\PLDIDIFF}[1]{#1}

\begin{document}

\title[\projectname: Accelerating Deep Neural Networks Execution with Advanced Operator Fusion]{\projectname: Accelerating Deep Neural Networks Execution with Advanced Operator Fusion}

\author{Wei Niu}
\affiliation{
  \institution{William \& Mary, USA}
  \country{}
}
\email{wniu@email.wm.edu}

\author{Jiexiong Guan}
\affiliation{
  \institution{William \& Mary, USA}
  \country{}
}
\email{jguan@email.wm.edu}

\author{Yanzhi Wang}
\affiliation{
  \institution{Northeastern University, USA}
  \country{}
}
\email{yanz.wang@northeastern.edu}

\author{Gagan Agrawal}
\affiliation{
  \institution{Augusta University, USA}
  \country{}
}
\email{gagrawal@augusta.edu}

\author{Bin Ren}
\affiliation{
  \institution{William \& Mary, USA}
  \country{}
}
\email{bren@wm.edu}


\begin{abstract} 

Deep Neural Networks (DNNs) have emerged as the core enabler of many major applications on mobile devices. To achieve high accuracy, DNN models have become increasingly deep with hundreds or even thousands of operator layers, leading to high memory and computational requirements for inference. Operator fusion (or kernel/layer fusion) is key optimization in many state-of-the-art DNN execution frameworks, such as TensorFlow, TVM, and MNN, that aim to improve the efficiency of the DNN inference. However, these frameworks usually adopt fusion approaches based on certain patterns that are too restrictive to cover the diversity of operators and layer connections, especially those seen in many extremely deep models. Polyhedral-based loop fusion techniques, on the other hand, work on a low-level view of the computation without operator-level information, and can also miss potential fusion opportunities. To address this challenge, this paper proposes a novel and extensive loop fusion framework called \projectname. The basic idea of this work is to work at an operator view of DNNs, but expand fusion opportunities by developing a classification of both individual operators and their combinations. In addition, \projectname includes 1) a novel mathematical-property-based graph rewriting framework to reduce evaluation costs and facilitate subsequent operator fusion, 2) an integrated fusion plan generation that leverages the high-level analysis and accurate light-weight profiling, and 3) additional optimizations during fusion code generation. \projectname is extensively evaluated on 15 DNN models with varied types of tasks, model sizes, and layer counts. The evaluation results demonstrate that \projectname finds up to $8.8 \times$ higher fusion opportunities, outperforms four state-of-the-art DNN execution frameworks with $9.3\times$ speedup. The memory requirement reduction and speedups can enable the execution of many of the target models on mobile devices and even make them part of a real-time application. 

\end{abstract}



\maketitle

\section{Introduction}

The past ten  years have witnessed a resurgence of Machine Learning, specifically in the form of Deep Learning. Deep Neural Networks (DNNs) such as Convolution Neural Networks (CNN) and Recurrent 
Neural Networks (RNN) serve as the state-of-the-art foundation and core enabler of many applications that have only emerged 
within the last few years  and yet have become extremely popular among all users of computing today ~\cite{rodgers2014recent,bhattacharya2016smart}. 
Behind the success of deep learning are the increasingly large model sizes and complex model structures that require tremendous computation and memory resources~\cite{dean2012large}. 
There is a difficult trade-off between increasing complexity of DNNs (required for increasing accuracy) and  deployment of these  DNNs  on resource-constrained 
mobile devices (required for wider reach).   

In recent years, there has been a significant emphasis on optimizing the execution of large DNNs. 
Operator fusion (or {\em kernel}/{\em layer} fusion)   has been a common approach towards improving  efficiency of DNN execution~\cite{chen2018tvm,TensorFlow-Lite}.   
The basic idea of such fusion is the same as the traditional loop fusion done by optimizing compilers~\cite{megiddo1997optimal,kennedy1993maximizing,kandemir1998improving}, and they 
lead to the following benefits:   (i) eliminating unnecessary materialization of intermediate results,  (ii) reducing unnecessary scans of the  input;  
and (iii) enabling other optimization opportunities.  
Traditional end-to-end frameworks like TensorFlow Lite~\cite{TensorFlow-Lite}, TVM~\cite{chen2018tvm}, MNN~\cite{Ali-MNN}, and Pytorch-Mobile ~\cite{Pytorch-Mobile} all have operator fusion optimizations, which 
are broadly based on recognizing certain fusion patterns.  These transformations have generally been based on a representation 
called {\em computational graph}~\cite{chen2018tvm}, which views the application as a set of operations on tensors, and representation 
of dependencies in the form of consumption of tensor(s) output by an operation by another operation. 

In this paper, we observe that the  fusion patterns considered in the past work~\cite{chen2018tvm,TensorFlow-Lite} are too restricted to cover the diversity  of  operators and 
layer connections that are emerging.  For example,  ONNX (Open Neural Network Exchange)~\cite{onnx-cite} lists 167 distinct operators,   and creating fusion patterns based on their combinations is unlikely to be a feasible approach.  
At the same time, traditional compiler loop transformations (including fusion~\cite{megiddo1997optimal,kennedy1993maximizing}) work on a low-level 
view of  the computation, i.e., loop (indices) and  dependence between array elements. More recent work on  
loop fusion has been based on polyhedral analysis~\cite{bondhugula2008practical}, with several different resulting algorithms~\cite{bondhugula2010model, acharya2018polyhedral, acharya2020effective}. 
Polyhedral analysis, while providing an excellent  foundation to rigorously reason the legality of, and explore the space of,  loop transformations, can be an ``overkill''  to capture 
the relatively simple data structures (tensors) and  operations (without loop-carried dependencies) in DNNs. Moreover,  polyhedral analysis is normally limited 
to affine-loop analysis and transformations (although latest efforts~\cite{venkat2014non, venkat2015loop,selva2019building} do extend it to certain non-affine loop optimizations), and cannot 
capture certain  operation (combinations) in DNNs. An  example  will  be  a 
combination of 
 {\tt  Gather}, which copies input to output indirectly using an  index array 
 followed by {\tt Flatten},  which changes the dimensionality of a tensor. 
 Finally, the operator view in computational graphs can enable us to exploit properties of  
these computations, which may be lost when a lower-level view of the computation is considered. 

\begingroup
\setlength{\tabcolsep}{1.5pt}
\begin{table}[t!]
\centering
\caption{{\bf An empirical study to motivate this work:} The relation of overall computation, layer count, and execution efficiency of multiple DNNs. Results are collected on Qualcomm Adreno 650 GPU with an optimized baseline framework with fixed-pattern operator fusion that outperforms all state-of-the-art DNN execution frameworks (called {\tt OurB+} and will be introduced later).}
\small{
\begin{tabular}{l|ccc|c}
    \toprule
    Model & \makecell{\#Total layer} & \makecell{IR size} & \#FLOPS & \makecell{Speed (FLOPs/S)}\\
    \hline
    VGG-16~\cite{simonyan2014very} & 51 & 161M & 31.0B & 320G \\
    YOLO-V4~\cite{bochkovskiy2020yolov4} & 398 & 329M & 34.6B & 135G \\
    DistilBERT~\cite{sanh2019distilbert} & 457 & 540M & 35.3B & 78G \\
    MobileBERT~\cite{sun2019mobilebert} & 2,387 & 744M & 17.6B & 44G \\
    GPT-2~\cite{radford2019language} & 2,533 & 1,389M & 69.1B & 62G \\
    \bottomrule
\end{tabular}
}
\label{tab:model_layer_comparison}
\end{table}
\endgroup

This paper presents \projectname, a rigorous and extensive loop fusion framework that can exploit the operator view of computations in DNNs, and yet can enable a set 
of advanced transformations.  The core idea is to  classify operators into different types, and develop rules for different combinations of the types, as opposed 
to looking for patterns with specific combination of operations.  Particularly, we 
first classify the existing operations in a DNN into several groups based on the mapping between their input and output , such as One-to-One, One-to-Many, and others.   
We also enhance  the computational graph representation into the Extended Computational Graph (ECG) representation, where the type (and other properties) of the operation 
are explicitly noted. 
Then,  we  design a mapping type analysis to infer  the profitability of fusing operations of different combinations of these types of operators, binning the combination into 
three groups: likely profitable (and legal), likely not profitable, and ones where profitability may need to be determined through profile information.   

Next, on the ECG representation, we apply a series of {\em graph rewriting rules} that we have developed. These rules exploit the mathematical properties 
of the operations and have a   similar flavor  to the  classical optimization  called {\em strength reduction}~\cite{cooper2001operator}. Unlike 
traditional compiler work, however, 
 we apply these rules on operations on tensors (and not scalars) and our set of rules go well beyond the traditional ones.  
The rest of our framework comprises 
algorithms for determining fusion of specific operations (based on certain heuristics) and generating optimized fused code.  
\PLDIDIFF{Almost each fusion generates a new operator (and its implementation) that is not present in the original library; however, once a new operator is generated, its implementation  can be reused when the same pattern is detected in the same or a different model.}
Overall, we show that an operator view of the DNN can enable rigorous optimizations, beyond what will be possible with a lower-level 
view of the computation or the existing (simplistic) work on applying a small set of fusion patterns on the operator view. 

\begingroup
\setlength{\tabcolsep}{4pt}
\begin{table*}[t]
\centering
\caption{{\bf Classification of DNN operators in mapping types.} These operators are defined in ONNX~\cite{onnx-cite}.}
\label{tab:layer_fusion_example}
\small
\begin{tabular}{l|l|c}
     \toprule
     {Mapping type} & Operators & Representative \\ \hline
     {One-to-One}   & \makecell[l]{Add, Asin, BatchNormalization, Cast, Ceil, Clip, Concat, Cos, Erf, Exp, Greater, LeakyRelu, Log, Not, \\PRelu, Reciprocal, Relu, Round, Sigmoid, Sin, Slice, Split, Sqrt, Tanh, Where}  & Add, Relu \\ \hline
     {One-to-Many}  & Elementwise w/ broadcast, Expand, Gather, Resize, Upsample  & Expand \\ \hline
     {Many-to-Many}  & \makecell[l]{AveragePool, CONV, ConvTranspose, CumSum, Einsum, GEMM, InstanceNormalization, MaxPool, \\Reduce (e.g. ReduceProd, ReduceMean), Softmax }  & Conv, GEMM \\ \hline
     {Reorganize}   & Flatten, Reshape, Squeeze, Unsqueeze  & Reshape \\ \hline
     {Shuffle}      & DepthToSpace, SpaceToDepth, Transpose  & Transpose \\ \bottomrule
\end{tabular}
\end{table*}
\endgroup

In summary, this paper makes the  following contributions:

\begin{itemize}[leftmargin=*,noitemsep,nolistsep]

\item It designs high-level abstractions (including mapping type analysis and ECG) for operator fusion by leveraging high-level DNN operator information. The approach 
can handle a diversity of operators and yet enable aggressive optimizations.

\item It proposes a novel mathematical-property-based graph rewriting to simplify ECG structure, optimize DNN computations, and facilitate subsequent fusion plan generation. 

\item It presents an integrated fusion plan generation by combining the benefit of efficient machine-independent mapping type analysis while leveraging 
a  profiling result database.

\item It implements optimized fusion code generation, integrating  the approach into a state-of-the-art end-to-end DNN execution framework. The optimized framework with operator fusion is called \projectname.  

\end{itemize}

\projectname is extensively evaluated on 
15 cutting-edge DNN models with 5 types of tasks, varied model sizes, and different layer counts on mobile devices.
Comparing with four popular state-of-the-art end-to-end DNN execution frameworks, MNN~\cite{Ali-MNN}, TVM~\cite{chen2018tvm}, TensorFlow-Lite~\cite{TensorFlow-Lite}, and Pytorch-Mobile~\cite{Pytorch-Mobile}, \projectname achieves up to  $8.8\times$ more loop fusions, $9.3\times$ speedup with our proposed advanced operator fusion. Particularly, \projectname {\em \ for the first time} allows many latest DNN models that are not supported by any existing end-to-end frameworks to run on mobile devices efficiently, even in real-time. 
Moreover, \projectname improves cache performance and device utilization -- thus, enabling execution on  devices with more restricted resources -- 
and reduces  performance tuning time during compilation. 






\section{Blessing  and Curse of Deep Layers}

This section presents a study that motivates our work, by demonstrating 
that it is challenging to execute  deep(er) neural networks efficiently, particularly on resource-constraint mobile  
devices, due to the  high  memory and computation requirements. 

As we stated earlier, there has been a trend towards deeper DNNs. With increasing 
amount of computation, there has also been a trend towards reducing the computation 
by reducing the weight size. 
Consider the well-known Natural Language Processing (NLP) model, BERT~\cite{devlin2018bert} as an example. TFLite takes 985ms to inference  BERT on the latest CPU of Snapdragon 865.
In recent efforts~\cite{sun2019mobilebert, radford2019language} (MobileBERT, GPT-2),  
machine learning researchers have addressed this issue by reducing the weight size on each layer and thus training thinner and deeper models to balance the computation workload and 
model accuracy.

However, we have observed that the depth of the model is the critical impediment to efficient execution. 
Our experimental study has correlated execution efficiency with the total amount of computation and the number of layers (Table~\ref{tab:model_layer_comparison}). 
Particularly, we can see  that although DistilBERT~\cite{sanh2019distilbert} and VGG-16~\cite{simonyan2014very} have a similar number of computations (while having 457 and 51 layers, respectively), 
DistilBERT's execution performance (78 GFLOPs/S) is much worse than VGG's (320 GFLOPs/S). This is mainly because of two reasons. 
First, models with more layers usually generate more intermediate results, thus increasing the memory/cache pressure. Second, deep models usually have an 
insufficient amount of computations in each layer, thus degrading the processor's utilization, particularly for  
GPUs.   Operator fusion can be an effective technique to reduce memory requirements and improve efficiency, and is the focus of our study.

\section{Classification of DNN Operators and Fusion Opportunity Analysis}\label{sec:fusion-classification}

This section establishes the basis for our approach, by classifying DNN operators and their combinations. 

\subsection{DNN Operators Classification}

This work carefully studied  all operators supported by a popular (and general) DNN ecosystem ONNX (Open Neural Network Exchange)~\cite{onnx-cite}, and finds that the mapping relation between (each) input  
and output of each operator is critical to determine both the  profitability and correct implementation  
of fusion optimization. Moreover, it is possible for us to classify all operators into five high-level abstract types based on the relationship between 
input elements and output elements.  These five types are One-to-One, One-to-Many, Many-to-Many (which includes Many-to-One, but we do not consider it separately 
here), Reorganize, and Shuffle.  
This classification  serves  as the  foundation of our proposed fusion framework. Table~\ref{tab:layer_fusion_example} shows more details of this operator classification and gives one or two representative examples for each mapping type. If an operator has only one input or multiple inputs with the same mapping type to the output, the mapping type of this operator is decided by its any input/output pair. If multiple input/output pairs with varied mapping types exist, this operator's mapping type is decided by the more complex mapping type\footnote{Order in  increasing order of complexity: One-to-One, Reorganize, Shuffle, One-to-Many, and Many-to-Many}.

Assuming each input element can be denoted as $x[d_1, \dots, \\d_n]$, \PLDIDIFF{where $x$ means the operand of an operator and $d_1, \dots ,d_n$ denotes the index for an element of an operand, } the mapping types between one input and one output are classified as follows.

\begin{itemize}[leftmargin=*,noitemsep,nolistsep]
\item {\bf One-to-One:}  There is  a set of functions ${F, f_1, \dots, f_n}$, such that  
$$y[d_1, \dots ,d_n] =  
F(x[f_1(d_1,), \dots ,f_n(d_n)])$$ 
and there is a 1-1 mapping between  each $[d_1, \dots ,d_n]$ and the corresponding  $[f_1(d_1,) \dots ,f_n(d_n)]$ used to 
compute it. 

\item {\bf One-to-Many:} There is a set of functions ${F, f_1, \dots, f_n}$, such that:  
$$y[e_1, \dots ,e_m]  = F(x[f_1(d_1), \dots ,f_n(d_n)])$$ 
where $m > n$, and there is a One-to-Many relationship between $[f_1(d_1), \dots ,f_n(d_n)]$ and 
$[e_1, \dots ,e_m]$. 

\item {\bf Many-to-Many:} There is  a set of functions $f^1_1, \dots ,f^1_n,  \dots $ \\ 
$f^k_1, \dots ,f^k_n$, such that:
$$y[e_1, \dots ,e_m] = F(x^1[f^1_1(d_1), \dots ,f^1_n(d_n)], \dots, $$ 
$$ \\x^k[f^k_1(d_1), \dots ,f^k_k(d_n)]).$$

\item {\bf Reorganize: } We have    
$$y[e_1, \dots ,e_m] = x[f_1(d_1), \dots ,f_n(d_n)]$$ 
and there is a 1-1 relationship 
between each $[e_1, \dots ,e_m]$ and the corresponding $[f_1(d_1), \dots ,f_n(d_n)]$.

\item {\bf Shuffle:} There is  a set of functions ${F, f_1,  \dots,  ,f_{n}}$, where $F$ is a {\em permutation} function, 
such that, 
$$y[e_1, \dots ,e_n] = x[f_1(d_{F(1)}), \dots ,f_n(d_{F(n)})].$$
\end{itemize}

 \begin{table}
\caption{{\bf Mapping type analysis.} The first  column  and the first row (both without color)  show the mapping types of first and second  operators, respectively, 
before fusion, and the colored cells show the mapping type of the operator after fusion. Green implies that these fusion combinations can be fused directly (i.e., they are profitable). 
Red  implies that these fusions are unprofitable.   Yellow implies that  further profiling is required to determine profitability. }
\label{tab:mapping-type-lookup}
\centering
\includegraphics[width=1\columnwidth]{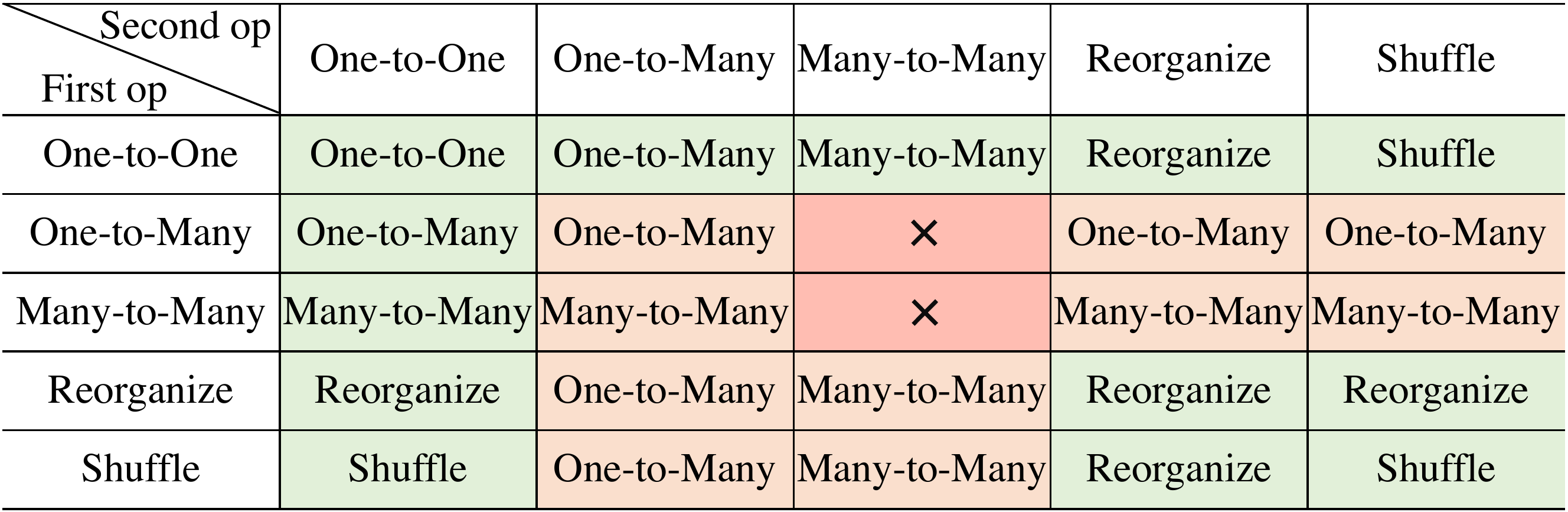}
\end{table}

\subsection{Fusion Opportunity Analysis}

Based on the mapping type of each operator, this work proposes a new fusion analysis. The basic idea is that given two fusion candidate operators with a certain combination
of mapping types, it is possible to: 1) infer the mapping type of the resulting  fused operation; and  2) simplify the profitability evaluation and correct 
implementation of this fusion. 

Table~\ref{tab:mapping-type-lookup} shows the details of this analysis. The first  column and the first row  (without any color) show the mapping types of the first 
and the second operator to be fused and the colored cells show the mapping type of the resulting operator. It further classifies the fusion of this combination of mapping types 
into three groups (shown as green, yellow, and red, respectively). Green implies that these fusions are legal and profitable and no further analysis is required. Red implies that these fusions are known to be  either illegal or clearly not profitable. Yellow implies that these fusions are legal; however, further profiling is required to determine profitability. 
This analysis eliminates the need for anytime runtime analysis or autotuning 
for red and green cases.  For remaining (yellow) cases, we can further  
accelerate compilation using a  {\em profiling database} 
that stores the execution results of various fusion combinations collected offline. 

 These five mapping types have a range of what we call  {\em transformation impedance} \PLDIDIFF{(which 
 we informally define as a metric to qualitatively express the difficulty to fuse)},  i.e., when they are fused with another type, they have 
different capability of deciding the fused mapping type. One-to-One has the lowest  transformation impedance  among all five types, 
whereas Reorganize and Shuffle's transformation  impedance is in the middle, i.e., they can transform One-to-One to their types while they 
cannot transform others. One-to-Many and Many-to-Many have the strongest transformation impedance, i.e., the resulted mapping type is decided by them solely when they are fused with other 
operators. 
Moreover, One-to-Many and Many-to-Many have the same capability, and Reorganize and Shuffle have the same as well. 

We elaborate on the following representative combinations to provide intuition behind the Table~\ref{tab:mapping-type-lookup}.

\begin{itemize}[leftmargin=*,noitemsep,nolistsep]
    \item {\em One-to-One with others.} When a One-to-One operator ($Op_1$ with the input {\tt I} and  the output {\tt O}) is fused with an operator of any 
    type ($Op_2$), i.e., $Op_2$ takes {\tt O} as the  input, the memory access to each element of {\tt O} can be mapped to the access to each element of {\tt I}, as long 
    as this mapping function is known. Unlike general programs  where the dependencies can be more complex, the use of tensors and a limited set of 
    operators limits the type of mappings,  and DNN operators carry this mapping information. Our analysis leverages this high-level operator information to ensure 
    the correctness of these fusions. Moreover, this fusion usually requires limited number of registers and does not incur extra overhead like data copying 
    or redundant computations, so they are profitable.    
    Take a case that fuses {\tt Add} and {\tt GEMM}  in either order.  Each element in the output of {\tt Add} can be replaced by two elements in the 
    two inputs of {\tt Add}, ensuring correct and profitable fusion, irrespective of the order of these operations. 

    \item {\em Reorder or Shuffle with others.} Both types are variants of One-to-One with a special mapping function  between the 
    input and  the output. Above reasons for the correctness analysis are also applied here; however, when fusing  with One-to-Many or Many-to-Many types operators, profitability 
    needs to be validated  with further profiling because of the possibility of  introduced data copying, change in data access order, 
    or redundant computations.  
    As an example, consider {\tt Expand} and {\tt Transpose} operators  -- \PLDIDIFF{{\tt Expand} copies the input tensor with a continuous memory access pattern}, whereas,  
    {\tt Transpose} transposes the input tensor to the output tensor according to the permutation in operator properties. Thus, the resulting fused operation may not have continuous memory accesses.

    \item {\em One-to-Many with Many-to-Many.} Take the  case that {\tt Expand} followed by {\tt Conv} -- as {\tt Conv} reads the feature map input tensor with continuous access, while a 
    One-to-Many operator can distribute the continuous input tensor elements.   
    As it is very desirable for the (compute-intensive) Many-to-Many operators to read the input tensors in a continuous way, we consider this fusion unprofitable. 
    
    \item {\em Many-to-Many with Many-to-Many.} When a Many-to-One mapping operator is followed by a Many-to-One operator, 
    e.g.  {\tt Conv} followed by another {\tt Conv}, attempting a combined execution will be too complicated and will likely negatively impact register and cache usage.  Thus, we consider them unprofitable. 
    
    \item {\em Many-to-Many with One-to-Many.} When a Many-to-One mapping operator is followed by a One-to-Many operator, 
    e.g.  {\tt Conv} followed by {\tt Expand} or {\tt Resize}, a combined execution may or may not have a desirable data access pattern. 
    When {\tt Conv} is combined with {\tt Expand}, as {\tt Expand} operator only expands a single dimension of the input, so it will not adversely affect the computation pattern of {\tt Conv}.  
    On the other hand, 
    if {\tt Conv} is combined with a {\tt Resize} that will copy the input tensor along  different dimensions, it can negatively impact  
    the computation of {\tt Conv}. Thus, we consider such cases to be requiring further profiling. 
   
\end{itemize}



\compactparagraph{Extended Computational Graph.} Based on the analysis above and as a background for the methods we will present next, we introduce Extended Computational Graph (ECG) as our intermediate representation (IR). 
As the name suggests, this represents builds on top of the (traditional) Computational Graph~\cite{chen2018tvm},  which 
captures the data-flow and basic operator information like the  operator type and parameters.    
ECG contains more fusion-related information, including {\tt mapping\_type} indicating the mapping type of each operator, {\tt IR\_removable} denoting if an intermediate result can be removed completely (which is true only if all its successors can be fused and which is calculated during fusion), and mathematical properties of the operations like 
whether the associative, commutative, and/or distributed properties hold.  

\section{\projectname's Design}

\subsection{Overview of \projectname}

\begin{figure}[t]
    \centering
    \includegraphics[width=\columnwidth]{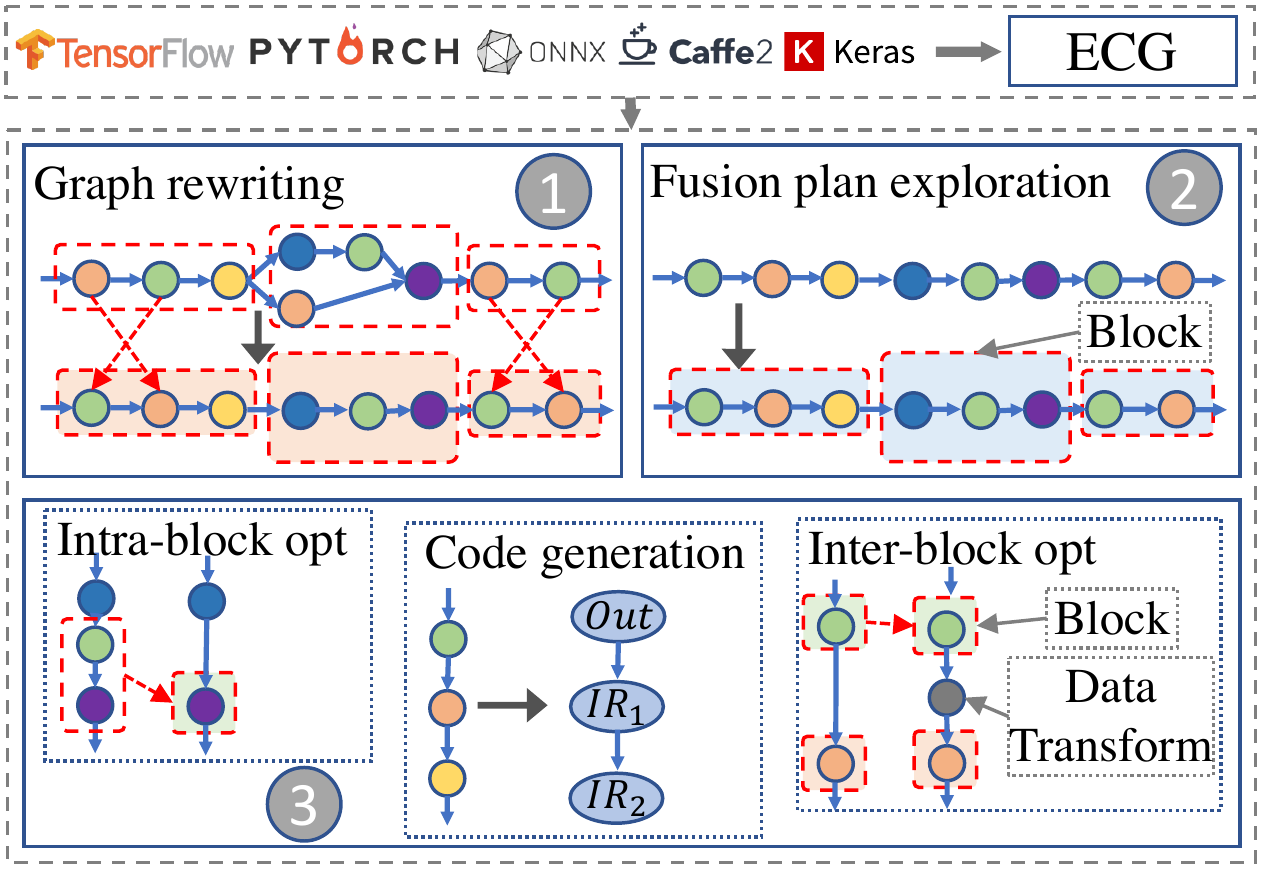}
    \caption{{\bf \projectname overview.}}
    \label{fig:system-overview}
\end{figure}

Figure~\ref{fig:system-overview} shows an overview of \projectname. 
It takes the computational graph generated from compiler-based DNN execution frameworks (e.g., TVM~\cite{chen2018tvm}, and MNN~\cite{Ali-MNN}) as  the 
input, and adds key information to create the Extended Computational Graph  (ECG). 
Based on this ECG, the main compiler optimization and code generation stage of \projectname consists of three  components:  \ding{202} mathematical-property-based graph rewriting (Section~\ref{sec:rewrite}), \ding{203}  lightweight profile-driven fusion plan exploration (Section~\ref{sec:plan}), and \ding{204} fusion code generation and other advanced fusion-based optimizations (Section~\ref{sec:code-gen}).



\begin{figure*}[t]
    \centering
    \includegraphics[width=0.96 \textwidth]{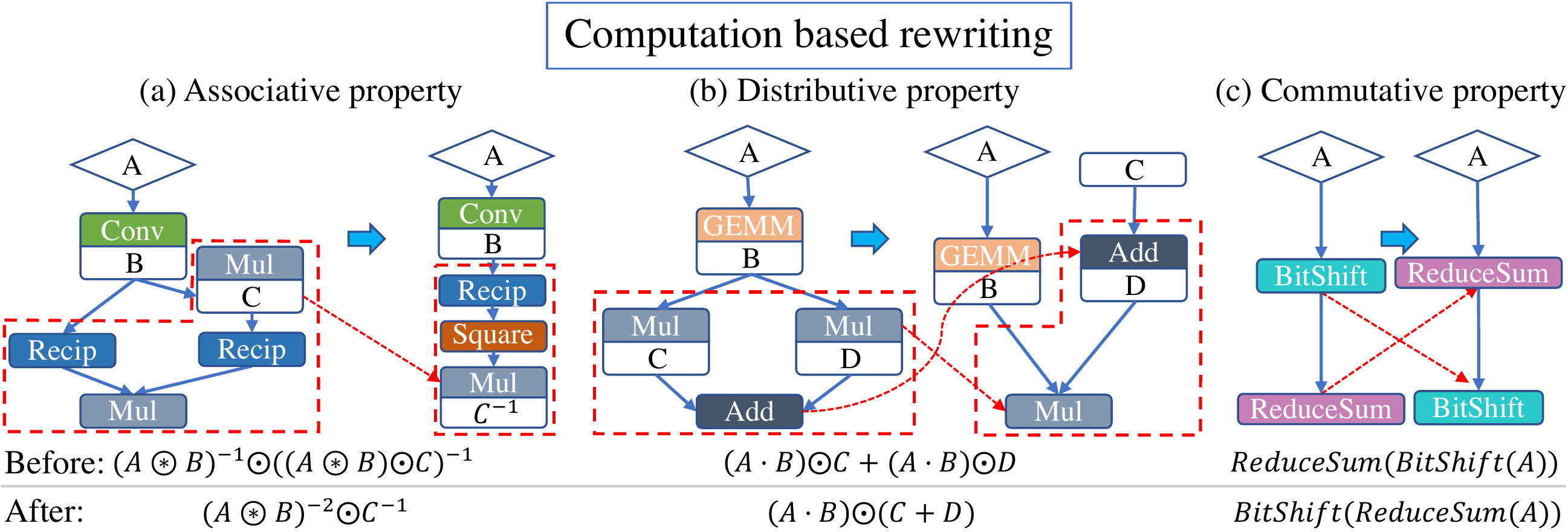}
    \caption{{\bf Examples of graph rewriting with mathematical properties.} Associative property
    explores the optimal execution order of operators and replaces the expensive combination of operators with a cheaper one.
    Distributive property
    explores the common combination of operators and simplifies the computation structure.
    Commutative property switches the execution order of operators to reduce the overall computation. 
    \PLDIDIFF{Note: the letter below each operator (e.g., {\tt B} below {\tt Conv} in (a)) or the letter in rectangle (e.g., {\tt C} in (b)) denotes that this input is from model weights rather than an intermediate result. The letter in diamond (e.g., {\tt A}) means that this is the input of this operator block, which could be the input of the model or intermediate result from a prior block. The intermediate results within this block are omitted for readability.}}
    \label{fig:fusion-type-example}
\end{figure*}

\begingroup
\setlength{\tabcolsep}{3.6pt}
\begin{table*}[t]
\centering
\caption{{\bf Graph rewriting  with mathematical properties.} \PLDIDIFF{Only representative graph rewriting rules are listed due to space limitation. In summary, \projectname derives 45, 38, and 66 graph rewriting rules in the category of Associative, Distributive, and Communicative, respectively.} We omit unrelated \PLDIDIFF{operators} for better readability. $\odot, +, -, Abs, Recip, Square, \sqrt{\phantom{x}}$ mean element-wise multiplication, addition, subtraction, absolute, reciprocal, square, and square root, respectively. {\tt BitShift} calculates the bit shifted value of elements of a given tensor element-wisely. {\tt ReduceSum} and {\tt ReduceProd} calculate the reduced summation and production of elements of an input tensor along an axis. {\tt Exp} calculates the exponent of elements in a given input tensor element-wisely. \#FLOPS denotes the number of floating point operations}
\label{tab:graph_rewrite_example}
\small
\begin{tabular}{l|P{6.0cm}r|P{5.9cm}r}
     \toprule
     \multirow{2}{*}{Property} & \multicolumn{2}{c|}{Without graph rewriting} & \multicolumn{2}{c}{With graph rewriting} \\ \cline{2-5}
     ~ &  Graph structure in equation & \#FLOPS & Graph structure in equation & \#FLOPS \\
     \hline
     \multirow{4}{*}{Associative} & $Recip(A) \odot Recip(A \odot B)$ & $4 * m * n$ & $Square(Recip(A)) \odot Recip(B)$ & $4 * m * n$ $\S$\\
     ~ & $(A \odot \sqrt{B}) \odot (\sqrt{B} \odot C)$ & $5 * m * n$ & $A \odot B \odot C$ & $2 * m * n$\\
     ~ & $Abs(A) \odot B \odot Abs(C)$ $^\dag$ & $4 * m * n$ & $Abs(A \odot C) \odot B$ & $3 * m * n$\\
     ~ & $(A \odot ReduceSum(B)) \odot (ReduceSum(B) \odot C)$ $^\P$ & $5 * m * n$ & $A \odot Square(ReduceSum(B)) \odot C$ & $3 * m * n + m$\\
     \hline
     \multirow{3}{*}{Distributive} & $A \odot C$ + $A \odot B$ & $3 * m * n$ & $(A + B) \odot C$ & $2 * m * n$\\
     ~ & $A + A \odot B$ & $2 * m * n$ & $A \odot (B + 1)$ & $2 * m * n$ $\S$ \\
     ~ & $Square(A + B) - (A + B) \odot C$ & $5 * m * n$ & $(A + B) \odot (A + B - C)$ & $3 * m * n$\\
     \hline
     \multirow{3}{*}{Commutative} & $A \odot B$  & $m * n$ & $B \odot A$ & $m * n$ $^\ddag$\\
     ~ & $ReduceSum(BitShift(A))$ $^\P$ & $2 * m * n$ & $BitShift(ReduceSum(A))$ & $m * n + m$\\
     ~ & $ReduceProd(Exp(A))$ $^\P$ & $2 * m * n$ & $Exp(ReduceSum(A))$ & $m * n + m$\\
     \bottomrule
     \multicolumn{5}{l}{{\scriptsize $^\S$ Although \#FLOPS is not reduced, A is loaded once instead of twice.}} \\
     \multicolumn{5}{l}{{\scriptsize $^\dag$ First use commutative property to swap $B$ and $Abs(C)$, then apply associative property.}} \\
     \multicolumn{5}{l}{{\scriptsize $^\ddag$ Even though this pattern has no \#FLOPS gains, it can enable further optimization, e.g the case of $\dag$.}} \\
     \multicolumn{5}{l}{{\scriptsize $^\P$ \#FLOPS is calculated by assuming the reduction of ReduceSum/ReduceProd is along with the inner-most dimension.}} \\
\end{tabular}
\end{table*}
\endgroup

\subsection{Mathematical-Property-Based Graph Rewriting}\label{sec:rewrite}

\projectname first employs a mathematical-property based graph rewriting pass to optimize the Extended Computational Graph (ECG). 
With this pass, \projectname is able to 1) remove unnecessary operations, 2) eliminate redundant intermediate data copies, and 3) replace costly (combination of) operators  with more efficient ones.  
This graph rewriting carried out here  is in the spirit of the classical compiler optimization of 
strength reduction~\cite{cooper2001operator}; however, here it is  performed   on  complicated operators on  matrices or tensors rather than  on 
scalar expressions. Moreover, the rules we present are more complex and involved, and are based on operations that are common in DNNs. 
More importantly, compared  to existing efforts on computational graph substitution (e.g., TASO~\cite{jia2019taso}), our  graph rewriting is designed to work in conjunction with operator  fusion and   identifies   a set of operators and rules  for that specific 
purpose.  Our evaluation results (Section~\ref{sec:eval}) show  that 
with graph rewriting, there are 18\% fewer fused layers left after fusion 
 on GPT-2. We also do an experimental comparison against 
 TASO later in this paper. 
 
Figure~\ref{fig:fusion-type-example} shows specific examples of  leveraged mathematical properties (distributive, communicative, and associative).
Table~\ref{tab:graph_rewrite_example} shows a more complete set of rules.  This table also shows the  computation size  (in \#FLOPS) before and after  the 
rewriting. Our rules  mainly focus on  operators in the  One-to-One mapping type (e.g., element-wise multiplication, addition, reciprocal, square root,  and 
others) and several reduction operators that are in Many-to-Many (e.g., {\tt ReduceSum} and {ReduceProd}) -- this is  because these operators usually follow our defined mathematical properties.   
\PLDIDIFF{\projectname uses \#FLOPs (rather than temporary output size or memory footprint) as the metric to drive 
graph rewriting mainly because of two reasons: first, in most of the applications scenarios of these rules, the temporary output size keeps the same before and after graph rewriting, and second, the size of the temporary output in a majority of other cases becomes a non-issue because fusion is applied after rewriting. For a small number of remaining cases, i.e., where temporary output size changes and the fusion is not applied, more sophisticated methods will be considered in the future.}

We now elaborate on some of the rules presented in Table~\ref{tab:graph_rewrite_example}, which were also depicted in Figure~\ref{fig:fusion-type-example}.

\renewcommand{\labelenumii}{\Roman{enumii}}
\begin{itemize}[leftmargin=*,noitemsep,nolistsep]
\item {\bf Associative:}
By leveraging the associative property, the graph rewriting pass can identify an optimized order of operators execution, and hence replace the expensive combination of operators with a new cheaper  
one.   Figure~\ref{fig:fusion-type-example} (a) shows an example, in which a combination of two {\tt Recip} operators and two {\tt Mul} operators is replaced by a combination of a {\tt Recip}, a {\tt Square}, and a {\tt Mul}. 
The latter is more efficient as it eliminates a {\tt Mul} operator and the intermediate result size is significantly reduced, leading to reduced register 
pressure after subsequent fusion. 

\item {\bf Distributive:}
Following the same ideas as above, applying distributive property also enables optimization opportunities. 
As shown in Figure~\ref{fig:fusion-type-example} (b), the combination of two {\tt Mul} operators and an {\tt Add} can be replaced by an {\tt Add} followed by a {\tt Mul}, thus 
 eliminating an unnecessary operator.

\item {\bf Commutative:}
The property guaranties the legality of swapping the position of two operators, which usually results in computation reduction. As shown in Figure~\ref{fig:fusion-type-example} (c), {\tt BitShift} \footnote{Calculate the bit shifted value of elements of a given tensor element-wisely.} and {\tt ReduceSum}\footnote{Calculate the reduced sum of elements of an input tensor along an axis.} satisfy communicative property, thus {\tt ReduceSum} can be scheduled to execute before {\tt BitShift}, reducing the number of elements on which {\tt BitShift} is applied. 
\end{itemize}

\projectname employs pattern matching~\cite{kounalis1991compilation, krebber2017non} to recognize rewriting candidates. 
However, associative and commutative matching is NP-complete~\cite{benanav1987complexity}.
Therefore, \projectname first partitions the entire Extended Computational Graph into many sub-graphs by considering operators with  neither of 
associative, communicative, or distributive properties as partitioning points within 
the original graph. 
\PLDIDIFF{
Within each sub-graph, \projectname can  explore all possible patterns and pattern combinations because these sub-graphs have limited number of operators. More specifically, all matching rules within a partition are considered and the rule leading to the largest reduction in \#FLOPS is applied. This process is repeated till there are no additional matching rules within the partition. \projectname chooses this greedy scheme to keep  the optimization overheads low.
}

\subsection{Light-Weight Profile-Driven Fusion Plan Exploration}\label{sec:plan}

\subsubsection{Overall Idea}

Optimal fusion plan generation  requires a large  search space~\cite{elgamal2017spoof, boehm2018optimizing}
and has been shown to be  NP-complete~\cite{kennedy1993maximizing, darte1999complexity}.
To keep the process at manageable costs,  \projectname  explores fusion plans by employing 
a new light-weight (greedy) approach based on our proposed Extended Computational Graph (ECG) IR and our classification of operations 
into mapping types.

The high-level ideas are as follows. 
First, \projectname selects the starting operators (called {\em fusion seed operators}) from our ECG  to restrict the search space. This is based on a key insight  
that operators of One-to-One mapping type have the potential  to yield more benefits because they  
a) potentially result in fusion of more layers, including both with their predecessors and successors because of what we refer 
to as lower transformation impedance, and 
b)  have lower memory  
requirements and need for fewer registers among all mapping types.  
Second, starting with these seed operators, \projectname explores fusion opportunities  along the seed operator's successors
and predecessors, respectively. 
Third, \projectname creates fusion plans based on an approach that 
combines machine-independent mapping type analysis and 
 a {\em  profiling result database}. The mapping type analysis follows Table~\ref{tab:mapping-type-lookup} to check the operators' mapping type combination (in ECG) to decide if these operators should be fused.  Such mapping  eliminates unnecessary profile data lookup for most cases.

\begin{figure}[t]
\vspace{0.2cm}
    \centering
    \includegraphics[width=\columnwidth]{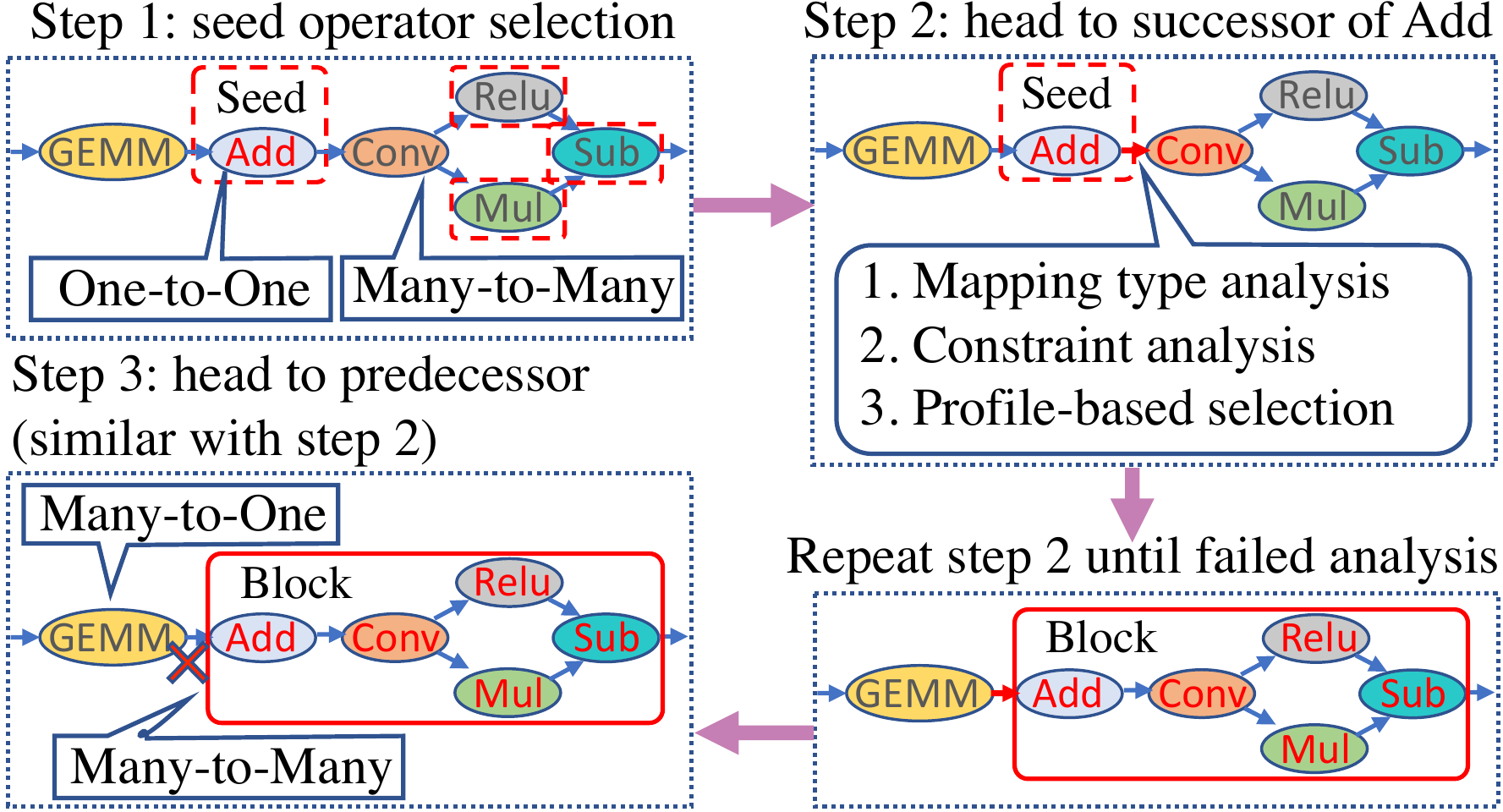}
    \caption{{\bf An example of fusion plan exploration.} Assume {\tt Add}, {\tt Conv}, {\tt Relu}, {\tt Mul}, and {\tt Sub} have identical output shape and IRS size.}
    \label{fig:fusion-plan-generation}
    \vspace{0.3cm}
\end{figure}

\subsubsection{Fusion Plan Generation Algorithm}

List~\ref{lst:fusion-plan-generation} shows our detailed fusion plan exploration algorithm. Its goal is to generate  candidate fusion blocks that  are  further optimized by subsequent intra-block optimizations (Section~\ref{sec:code-gen}) before fusion code generation. Figure~\ref{fig:fusion-plan-generation} illustrates its basic idea with a simplified example. This algorithm consists of three main steps:

\compactparagraph{Step I: Fusion seed operator(s)  selection.} \projectname selects the One-to-One operator with the minimum intermediate result as the fusion seed operator  (as shown in Listing~\ref{lst:fusion-plan-generation} lines 1 to 5). This heuristic is used because a smaller intermediate result makes fusion more profitable. 
\PLDIDIFF{This  may seem  counter-intuitive because fusing the operators with larger intermediate results usually results in more benefits. However, \projectname has a different goal, i.e., to ultimately enable more fusions to occur. Starting with a pair of operators with smaller intermediate results creates opportunities to fuse more (smaller) operators together, increase overall computation granularity, and hence enable higher parallelism and better load balance for the  entire  DNN computation. }
If multiple seed operators with the same minimum size of intermediate results exist, \projectname initiates fusion with them one after another 
(unless another seed is grouped to the same candidate fusion block). 
 In Figure ~\ref{fig:fusion-plan-generation}, {\tt Add}, {\tt Relu}, {\tt Mul}, and {\tt Sub} are in One-to-One type (with an identical intermediate result size), then {\tt Add} is selected as the seed for the first round of fusion plan exploration.

\lstset{numbers=left, numberstyle=\tiny, stepnumber=1, numbersep=3pt, frame=tb, caption=Fusion plan generation,emph={IR_deletable},emphstyle=\underbar,xleftmargin=2.5ex}
\begin{lstlisting}[label={lst:fusion-plan-generation},language=python]
def generate_seed(ops):
  # find all one_to_one mapping operators
  oto_ops = find_all_one_to_one(ops)
  # find the operator with minimum IRS size
  return min(op.IRS_size for op in oto_ops)

def fuse_successor(op, successor, block):
  # Step 2.1: check the mapping relationship
  relation = mapping_check(op, successor)
  # return if successor can not be fused
  if relation == fuse_break: return
  # Step 2.2: check the constraint requirement
  if not check_constraint(op, successor, block): 
    return
  # fuse by profile-based selection
  if relation == fuse_depend: 
    # Step 2.3: get latency w/ database/runtime
    temp_latency = latency(block + successor)
    if temp_latency > latency(block, successor):
        return
  block = op + successor
  # Step 2.4: recursively head to successor
  for fusing_op in successors(successor):
    fuse_successor(successor, fusing_op, block)

# Similar with fuse_successor
def fuse_predecessor(op, predecessor, block):
  # Similar with step 2.1, 2.2, 2.3, 2.4

# <Algorithm Entry>
unfused_ops = all_operators
# Step 1: start fuse from the selected seed
while(sp = generate_seed(unfused_ops))
  block = [sp]
  # Step 2: head to successor
  for successor in successors(sp):
    fuse_successor(sp, successor, block)
  # Step 3: head to predecessor
  for predecessor in predecessors(sp):
    fuse_predecessor(sp, predecessor, block)
  unfused_ops = unfused_ops - block     
\end{lstlisting}

\compactparagraph{Step II: Propagated exploration  along seed's successors.} 
Each operator may have one or multiple immediate  predecessors and successors. \projectname first processes the seed operator's  successors one by one (Listing ~\ref{lst:fusion-plan-generation} Lines 7 to 24). At any stage in this recursive exploration, if a node cannot be fused with any of its immediate successors, fusion is not considered 
any further. 
Broadly, this step proceeds as follows.  First, {\em mapping type analysis} 
(Listing ~\ref{lst:fusion-plan-generation} Step 2.1) categorizes the potential fusion result into three types based on Table~\ref{tab:mapping-type-lookup}: 
1) {\tt fuse\_break} indicates this is a Red case, and fusion should be aborted; 2) {\tt fuse\_through} indicates that this is a Green case, and should be proceeded without any further analysis; 3) {\tt fuse\_depend} indicates that this is a Yellow case, requiring a profile data lookup. 
Second,  
a {\em constraints check}  (Listing ~\ref{lst:fusion-plan-generation} Step 2.2) is applied 
to analyze if further fusion is likely undesirable, i.e, it  incurs too many overheads (e.g., can cause excessive   register spills).   
Using an empirically determined threshold, the algorithm can decide to not consider any additional fusion candidates. 
Otherwise,  \projectname continues exploring fusion candidates recursively.   Figure~\ref{fig:fusion-plan-generation} shows an example of fusing {\tt Add} with {\tt Conv} and  other 
operators  with this step. After this step, the generated candidate fusion block has a  mapping type of Many-to-One, and includes five operators 
({\tt Add}, {\tt Conv}, {\tt Relu}, {\tt Mul}, and {\tt Sub}).

\compactparagraph{Step III: Propagated exploration  along seed's predecessors.} After processing along the seed's successor direction, \projectname processes along the seed's predecessors direction with the same algorithm as Step II (In fact, Step III and Step II can be swapped).  However, one difference is that if an operator   has multiple immediate predecessors,  
there is an option of fusing with some, but not all, of these immediate predecessors. 
In the example in Figure~\ref{fig:fusion-plan-generation}, the first attempt of fusing current candidate fusion block with {\tt GEMM}  fails because both of them are  of 
many-to-one mapping type. Table~\ref{tab:mapping-type-lookup} indicates this is a {\tt fuse\_break} case, so {\tt GEMM} is not included in this candidate fusion block.  

\compactparagraph{Iterate.} \projectname completes a round of fusion plan generation with above steps. If more fusion seeds exist, it will iterate from Step II with the next seed until  
no additional fusion seed is available.

\begin{figure}[t!]
    \centering
    \includegraphics[width=\columnwidth]{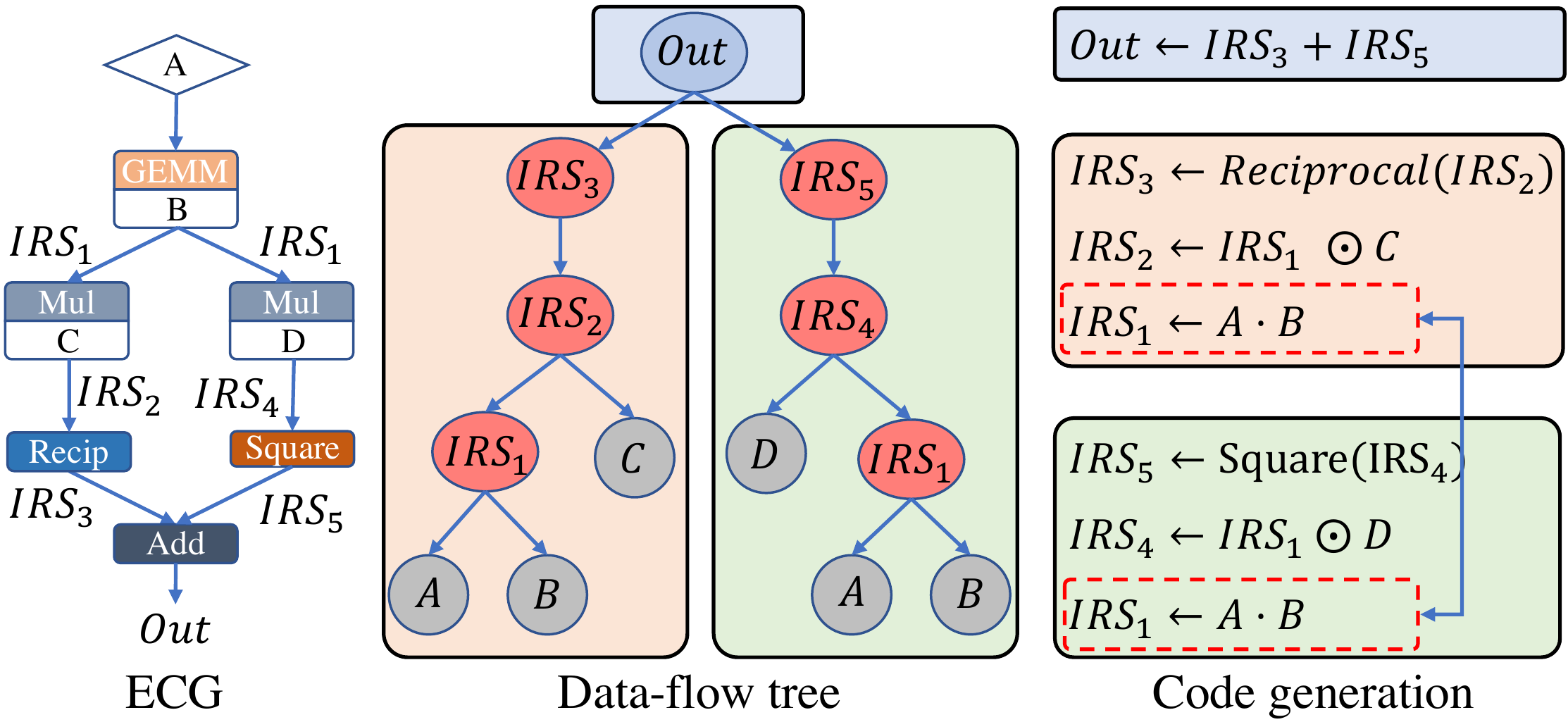}
    \caption{{\bf Code generation.}}
    \label{fig:code-generation-tree}
\end{figure}

\subsection{Fusion Code Generation and Optimizations}\label{sec:code-gen}

\subsubsection{Fusion Code Generation}

\PLDIDIFF{
Once fusion blocks have been selected by our algorithm,  \projectname generates fused code for each fusion block with a {\em data-flow tree} (DFT) built from the Extended Computational Graph (ECG) and a set of pre-defined code generation rules. 
\projectname generates C++ code for mobile CPU and OpenCL for mobile GPU, respectively.
More specifically, \projectname traverses DFT and generates fused code for each pair of operators to be fused 
by leveraging the code generation rules that are based on abstract mapping types (e.g., One-to-One).
 23 code generation rules are defined for each of mobile CPU and mobile GPU, with one rule corresponding to a green or yellow cell in Table ~\ref{tab:mapping-type-lookup}. The basic idea is that as long as the operators are of the same type, the same rules lead to efficient code. While fusing more than two operators, these rules are invoked each time two operators are fused. Finally, the subsequent code optimizations (e.g, vectorization, unrolling, tiling, and memory/register optimizations, and auto-tuning of these optimizations) are handled by our existing framework called PatDNN~\cite{niu2020patdnn}, thus not a major contribution of this paper. 
 Note that almost each fusion generates a new operator (and its codes) that is not present in the original operator library; however, once the new operator (and its code) is generated, it can be used for both the current model and future models.
}

\PLDIDIFF{
Figure~\ref{fig:code-generation-tree} shows an example of the code generation.}
\PLDIDIFF{ To elaborate, 
\projectname first generates a data-flow tree (DFT) from the Extended Computational Graph (ECG). 
This DFT represents the final output ({\tt Out}), all intermediate results ({\tt IRS}), and all inputs ({\tt A}, {\tt B}, {\tt C}, and {\tt D}) with the edges 
 reversed as compared to the ECG (i.e., the parent node depends on the children nodes). 
 During the fused code generation, \projectname traverses this DFT 
 to recognize the input/output data dependence (and fuses corresponding ECG operations), recursively. 
The right-hand side of Figure~\ref{fig:code-generation-tree}, shows an example of this DFT traversal (the fused code generation based on the pre-defined code generation rules is omitted in this Figure for readability and is introduced in the next paragraph). First, \projectname recognizes that $Out$ depends on $IRS_3 \ +\ IRS_5$; next, it recognizes that $IRS_3$ depends on reciprocal of $IRS_2$, and so on, until reaching the input of $A, B, C, D$. It is worth noting \projectname can also find {\em redundant computations} in DFT with a common sub-tree identification and eliminate them during code generation. In our example, both {\tt Mul} operators use $IRS_1$, resulting in a common sub-tree in DFT, so the recognition in two red boxes of 
Figure~\ref{fig:code-generation-tree} is only taken once.}

\PLDIDIFF{
During this DFT traversal, \projectname employs the pre-defined code generation rules to generate the code for each pair of operators to be fused.
For the example shown in Figure~\ref{fig:code-generation-tree},
 \projectname first fuses {\tt Add} with its left input branch {\tt Recip}. Both {\tt Add} and {\tt Recip} belong to One-to-One mapping, and hence the fused operator is also One-to-One. \projectname keeps fusing {\tt Mul} (One-to-One) with this newly fused operator, and the result is still One-to-One. Next, this newly generated operator is fused with  {\tt GEMM} (Many-to-One), generating a new Many-to-One operator.
 Similar steps are taken along the right input branch of {\tt Add} until all operators are fused into a single new Many-to-One operator.
 \projectname relies on the DFT traversal introduced in the prior paragraph to figure out the input/output data dependence, and employs the operator mapping type to handle the index mapping relationship and generate proper nested loop structures.
 }
 
 \PLDIDIFF{
 To explain this further, here is an example with more complicated mapping types: {\tt GEMM} (Many-to-Many) + {\tt Div} (One-to-One) + {\tt Transpose} (Shuffle). First, \projectname fuses {\tt Transpose} and {\tt Div}, a case of (``Shuffle + One-to-One'') by first permuting the loop in  the {\tt Transpose} operator  and then fusing it  with the {\tt Div} operator. It generates a new operator  
 of the type  Shuffle. Next, \projectname fuses GEMM (Many-to-Many type) with this new operator (Shuffle type), in which \projectname maps output elements of GEMM to the destination that is decided by this new operator.  
 }

\subsubsection{Other Fusion-related Optimizations} \label{sec:other-opt}

\projectname also includes several advanced optimizations enabled by our fusion analysis and fused code generation. They broadly can be characterized into 
two groups, {\em intra-fusion-block optimizations}  that are performed on Extended Computational Graph (ECG) immediately before the code generation  and  
{\em inter-fusion-block optimizations}  on the generated fused code.

\begin{figure}[t]
    \centering
    \includegraphics[width=\columnwidth]{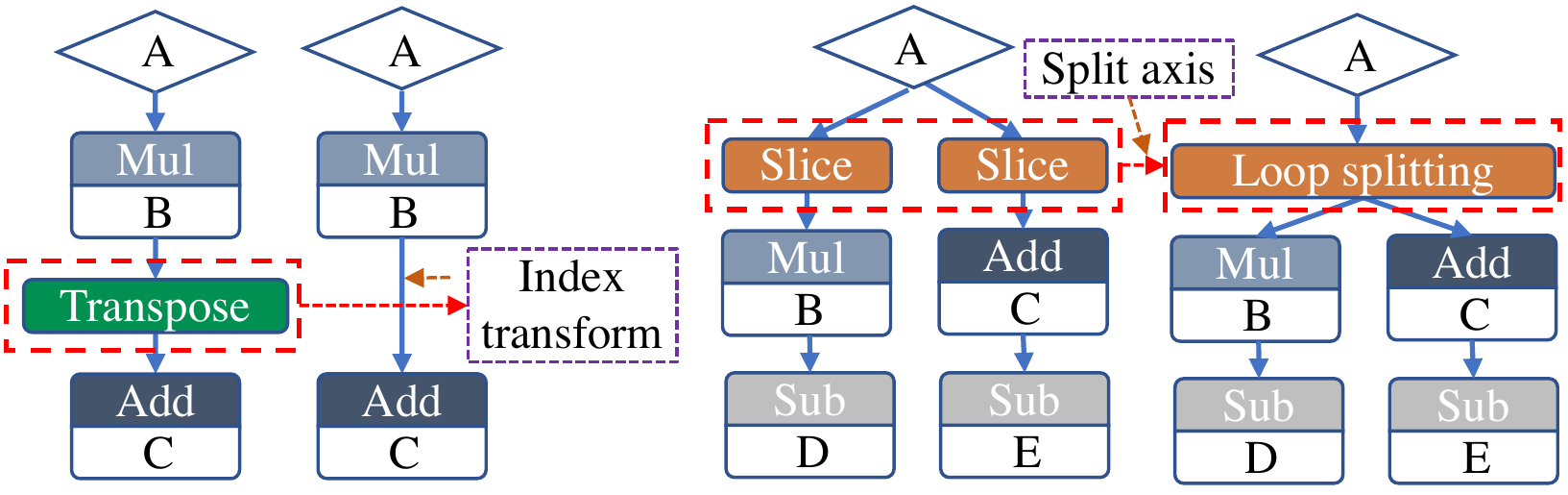}
    \caption{{\bf Data movement operators optimization.}}
    \label{fig:data-type-example}
\end{figure}

\compactparagraph{Intra-block Optimizations:} Operators in Shuffle and Reorganize mapping types usually  involve intensive data movement. 
We observed many of these time/memory consuming data operations can be eliminated.  In particular, consider the case when the transformed data 
is used by only one subsequent operator because the data locality improvement brought this data transformation cannot be compensated by the overhead of intermediate results generation and storage. Figure~\ref{fig:data-type-example} shows such  examples -- particularly, in these,  data transpose and data slicing operations bring more overheads than  
the benefit.  Thus, in such cases,  \projectname replaces them with 
operations that have a changed data index.   
These optimizations are performed after  graph rewriting and result in an ECG that should have a more efficient implementation. 

\compactparagraph{Inter-block Optimization:} Different operators prefer different data formats. Without the proposed graph rewriting optimizations and operator fusion, normally  such choices are made at the level of  each individual operator -- however, this 
can  result  in redundant or unnecessary transformations. In contrast, \projectname considers the data format choice at a global level, thus avoiding redundant or unnecessary transformations.
\PLDIDIFF{
Currently, \projectname employs a heuristic approach to optimize the data format, which is  as follows. For a specific fusion block, it identifies one {\em dominant} operator whose performance is impacted the most by the choice of the layout (e.g., {\tt CONV}, {\tt GEMM}, and {\tt Softmax} are most likely to such operators). The optimal layout for this operation is then used for the entire fusion block. This heuristic approach works based on a key observation that most other {\em non-dominant} operators can employ any layout/format without their performance being significantly affected.
A potential future work will be to  consider more sophisticated cost models, including 
balancing the cost of reformatting the data with reductions in  execution because of the optimized layout. 
}



\begingroup
\setlength{\tabcolsep}{1.3pt}
\begin{table*}[t!]
\centering
\caption{{\bf Fusion rate evaluation: computation layer count and intermediate result size for all evaluated DNNs.} CIL (Compute-Intensive Layer): each input is used more than once, e.g. MatMul, CONV. MIL (Memory-Intensive Layer): each input is used only once, e.g. Activation. \PLDIDIFF{IRS}: intermediate results. '-' means this framework does not support this model.}
\label{tab:eva_model_layer}
\small
\begin{tabular}{lcc|cccc|cccccr}
    \toprule
    \multirow{2}{*}{Model} & \multirow{2}{*}{Type} & \multirow{2}{*}{Task} & \multicolumn{4}{c|}{Layer counts and \PLDIDIFF{IRS} sizes before opt.} & \multicolumn{6}{c}{Layer counts and \PLDIDIFF{IRS} sizes after opt.}\\ \cline{4-13}
    ~ & ~ & ~ & \#CIL & \#MIL & \#Total layer & IRS size & MNN & TVM & TFLite & Pytorch & {\bf DNNF} & {\bf IRS size}\\
    \hline
    EfficientNet-B0    & 2D CNN      & Image classification & 82  & 227   & 309   & 108MB   & 199 & 195          & 201   & 210 & {\bf 97}  & {\bf 26MB}  \\
    VGG-16             & 2D CNN      & Image classification & 16  & 35    & 51    & 161MB   & 22  & 22           & 22    & 22  & {\bf 17}  & {\bf 52MB}  \\
    MobileNetV1-SSD    & 2D CNN      & Object detection     & 16  & 48    & 202   & 110MB   & 138 & 124          & 138   & 148 & {\bf 71}  & {\bf 37MB}  \\
    YOLO-V4            & 2D CNN      & Object detection     & 106 & 292   & 398   & 329MB   & 198 & 192          & 198   & 232 & {\bf 135} & {\bf 205MB} \\
    C3D                & 3D CNN      & Action recognition   & 11  & 16    & 27    & 195MB   & 27  & 27           & -     & 27  & {\bf 16}  & {\bf 90MB}  \\
    S3D                & 3D CNN      & Action recognition   & 77  & 195   & 272   & 996MB   & -   & -            & -     & 272 & {\bf 98}  & {\bf 356MB} \\
    U-Net              & 2D CNN      & Image segmentation   & 44  & 248   & 292   & 312MB   & 241 & 232          & 234   & -   & {\bf 82}  & {\bf 158MB} \\
    Faster R-CNN       & R-CNN       & Image segmentation   & 177 & 3,463 & 3,640 & 914MB   & -   & -            & -     & -   & {\bf 942} & {\bf 374MB} \\
    Mask R-CNN         & R-CNN       & Image segmentation   & 187 & 3,812 & 3,999 & 1,524MB & -   & -            & -     & -   & {\bf 981} & {\bf 543MB} \\
    TinyBERT           & Transformer & NLP                  & 37  & 329   & 366   & 183MB   & -   & 304$^\dag$   & 322   & -   & {\bf 74}  & {\bf 55MB}  \\
    DistilBERT         & Transformer & NLP                  & 55  & 402   & 457   & 540MB   & -   & 416$^\dag$   & 431   & -   & {\bf 109} & {\bf 197MB} \\
    ALBERT             & Transformer & NLP                  & 98  & 838   & 936   & 1,260MB & -   & 746$^\dag$   & 855   & -   & {\bf 225} & {\bf 320MB} \\
    BERT$_\text{BASE}$ & Transformer & NLP                  & 109 & 867   & 976   & 915MB   & -   & 760$^\dag$   & 873   & -   & {\bf 216} & {\bf 196MB} \\
    MobileBERT         & Transformer & NLP                  & 434 & 1,953 & 2,387 & 744MB   & -   & 1,678$^\dag$ & 2,128 & -   & {\bf 510} & {\bf 255MB} \\
    GPT-2              & Transformer & NLP                  & 84  & 2,449 & 2,533 & 1,389MB & -   & 2,047$^\dag$ & 2,223 & -   & {\bf 254} & {\bf 356MB} \\
    \bottomrule
    \multicolumn{13}{l}{\makecell[l]{{\scriptsize $^\dag$ TVM does not support this model on mobile. This layer count number is collected on a laptop platform for reference.}}} \\
\end{tabular}
\end{table*}
\endgroup

\section{Evaluation}\label{sec:eval}
\PLDIDIFF{
 \projectname is implemented on top of  an existing end-to-end DNN execution framework called PatDNN~\cite{niu2020patdnn} that supports both dense and sparse DNN execution.} It has been shown in our previous work that  
 PatDNN~\cite{niu2020patdnn} performs 
 slightly better than TVM, MNN, and TFLITE  even without our proposed operator fusion.  
  For readability, we also call this optimized framework \projectname. 
 Our evaluation has four objectives: 
 1) demonstrate that the proposed fusion framework (together with graph rewriting) is effective by showing how 
\projectname outperforms other state-of-the-art frameworks, and no-fusion and fixed-pattern fusion implementations on  various DNN models; 
2) validating \projectname's generality by showing its efficient execution on both CPU and GPU 
on a wide spectrum of DNNs (for 5 types of tasks, with varied sizes, and layer counts ranging  from relatively shallow to extremely deep);  
3) analyzing the impact of different 
compiler optimizations on both  execution time and compilation time; 
and 4) demonstrating the effective  portability of \projectname by evaluating it on three different mobile phones.

More specifically,  \projectname (also called {\tt DNNF} for short)  is compared   against  four popular state-of-the-art end-to-end DNN execution frameworks: MNN~\cite{Ali-MNN}, TVM~\cite{chen2018tvm}, TensorFlow-Lite ({\tt TFLite})~\cite{TensorFlow-Lite}, and Pytorch-Mobile ({\tt Pytorch})~\cite{Pytorch-Mobile}. Because certain  extremely deep neural networks are not supported by any of these existing frameworks (or just supported by their mobile CPU implementation), we also set a baseline by turning off \projectname's all fusion related optimizations (called {\tt OurB}, i.e., our baseline version without fusion) and implement a version that optimizes {\tt OurB} with fixed-pattern fusion (using operator fusion described in TVM~\cite{chen2018tvm}) (called {\tt OurB+}), and compare \projectname against them.

\subsection{Evaluation Setup}

\compactparagraph{Models and datasets.} 
\projectname is evaluated on 15 mainstream DNN models. Table~\ref{tab:eva_model_layer} characterizes them with a comparison of their targeted task,  number 
of parameters, total number of  layers, and number of floating point operations (FLOPS). Particularly, we have
1) two image classification 2D CNNs (EfficientNet-B0 ~\cite{tan2019efficientnet} and VGG-16 ~\cite{simonyan2014very}),  
2) two object detection  two-dimensional (2D) CNNs (MobileNetV1-SSD ~\cite{Liussd2016}  and YOLO-V4 ~\cite{bochkovskiy2020yolov4}),  
3) two  action recognition three-dimensional (3D)  CNNs (C3D ~\cite{tran2015learning}  and S3D ~\cite{xie2018rethinking}), 
4) one image segmentation 2D CNN (U-Net ~\cite{ronneberger2015u}) and two image segmentation R-CNNs (Mask R-CNN ~\cite{he2017mask}  and FasterRCNN ~\cite{ren2015faster}), and 
5) six natural language processing (NLP) models (TinyBERT ~\cite{jiao2019tinybert}, DistilBERT ~\cite{sanh2019distilbert}, ALBERT ~\cite{Lan2020ALBERT}, BERT$_\text{BASE}$, MobileBERT ~\cite{sun2019mobilebert}, and GPT-2 ~\cite{radford2019language}). 

\begingroup
\setlength{\tabcolsep}{3.2pt}
\renewcommand\arraystretch{1.0}
\begin{table*}[t!]
\centering
\caption{{\bf Inference latency comparison: \projectname, MNN, TVM, TFlite, and PyTorch on mobile CPU and GPU.} \#FLOPS denotes the number of floating point operations. OurB is our baseline implementation by turning off all fusion optimizations and OurB+ is OurB with a fixed-pattern fusion as TVM. DNNF is short for \projectname, i.e., our optimized version. '-' denotes this framework does not support this execution.}
\label{tab:eva_performance_report}
\small
\begin{tabular}{l|cc|cc|cc|cc|cc|cc|cc|cr}
     \toprule
     \multirow{2}{*}{Model} & \multirow{2}{*}{\#Params} & \multirow{2}{*}{\#FLOPS} & \multicolumn{2}{c|}{MNN (ms)} & \multicolumn{2}{c|}{TVM (ms)} & \multicolumn{2}{c|}{TFLite (ms)} & \multicolumn{2}{c|}{Pytorch (ms)} & \multicolumn{2}{c|}{OurB (ms)} & \multicolumn{2}{c|}{OurB+ (ms)} & \multicolumn{2}{c}{{\bf DNNF (ms)}} \\ \cline{4-17}
     ~                  & ~ & ~  & CPU & GPU & CPU & GPU & CPU & GPU   & CPU   & GPU & CPU & GPU & CPU & GPU & {\bf CPU} & {\bf GPU} \\ \hline
     EfficientNet-B0    & 5.3M & 0.8B  & 41 & 26 & 56 & 27 & 52  & 30 & 76 & - & 54 & 35 & 38 & 24 & {\bf 16} & {\bf 10}  \\
     VGG-16             & 138M & 31.0B & 242 & 109 & 260 & 127 & 245  & 102 & 273 & - & 251 & 121 & 231 & 97 & {\bf 171} & {\bf 65}  \\
     MobileNetV1-SSD    & 9.5M &  3.0B & 67 & 43 & 74 & 52 & 87  & 68 & 92 & - & 79 & 56 & 61 & 39 & {\bf 33} & {\bf 17} \\
     YOLO-V4            &  64M & 34.6B & 501 & 290 & 549 & 350 & 560  & 288 & 872 & - & 633 & 390 & 426 & 257 & {\bf 235} & {\bf 117} \\
     C3D                &  78M & 77.0B & 867 & - & 1,487 & - & - & - & 2,541 & - & 880 & 551 & 802 & 488 & {\bf 582} & {\bf 301} \\
     S3D                & 8.0M & 79.6B & - & - & - & - & - & - & 6,612 & - & 1,409 & 972 & 1,279& 705 & {\bf 710} & {\bf 324} \\
     U-Net              & 2.1M & 15.0B & 181 & 106  & 210 & 120 & 302 & 117 & 271 & - & 227 & 142 & 168 & 92  & {\bf 99}  & {\bf 52}  \\
     Faster R-CNN       &  41M & 47.0B & - & - & - & - & - & - & - & - & 2,325 & 3,054 & 1,462 & 1,974 & {\bf 862} & {\bf 531} \\
     Mask R-CNN         &  44M &  184B & - & - & - & - & - & - & - & - & 5,539 & 6,483 & 3,907 & 4,768 & {\bf 2,471} & {\bf 1,680} \\
     TinyBERT           &  15M &  4.1B & - & - & - & - & 97 & - & - & - & 114 & 89 & 92 & 65 & {\bf 51} & {\bf 30} \\
     DistilBERT         &  66M & 35.5B & - & - & - & - & 510 & - & - & - & 573 & 504 & 467 & 457 & {\bf 224} & {\bf 148} \\
     ALBERT             &  83M & 65.7B & - & - & - & - & 974 & - & - & - & 1,033 & 1,178 & 923 & 973 & {\bf 386} & {\bf 312} \\
     BERT$_\text{Base}$ & 108M & 67.3B & - & - & - & - & 985 & - & - & - & 1,086 & 1,204 & 948 & 1,012 & {\bf 394} & {\bf 293} \\
     MobileBERT         &  25M & 17.6B & - & - & - & - & 342 & - & - & - & 448 & 563 & 326 & 397 & {\bf 170} & {\bf 102} \\
     GPT-2              & 125M & 69.1B & - & - & - & - & 1,102 & - & - & - & 1,350 & 1,467 & 990 & 1,106 & {\bf 394} & {\bf 292} \\
     \bottomrule
\end{tabular}
\end{table*}
\endgroup

Because the choice of datasets has a negligible impact on the final inference latency or relative execution speeds (and also because of space limitations), 
we report  results from one dataset for each model. 
EfficientNet-B0 and VGG-16 are trained on ImageNet dataset~\cite{deng2009imagenet}; MobileNetV1-SSD and YOLO-V4 are trained on MS COCO~\cite{lin2014microsoft}; C3D and S3D are trained on UCF-101~\cite{soomro2012ucf101}; U-Net, Faster R-CNN, and Mask R-CNN are trained on PASCAL VOC 2007~\cite{everingham2007pascal}; TinyBERT, DistilBERT, ALBERT, BERT$_{\text{base}}$, MobileBERT, and GPT-2 are trained on BooksCorpus ~\cite{devlin2018bert} and English Wikipedia~\cite{devlin2018bert}. 
Because the  model accuracy is identical among all frameworks, and also because of space limitations, we only focus on execution times 
and do not report accuracy. 

\compactparagraph{Evaluation environment.}
The evaluations are carried out  on a  Samsung Galaxy S20 cell phone  that has  Snapdragon 865 processor \PLDIDIFF{~\cite{snapdragon-865}}, 
which comprises  an octa-cores Kryo 585 CPU and Qualcomm Adreno 650 GPU yielding high performance with good power efficiency. For demonstrating portability, 
we further use a Samsung Galaxy S10 with a Snapdragon 855 \PLDIDIFF{~\cite{snapdragon-855}} (Qualcomm Kryo 485 Octa-core CPU and a Qualcomm Adreno 640 GPU), and an 
Honor Magic 2 with a  Kirin 980  \PLDIDIFF{~\cite{kirin-980}} (ARM Octa-core CPU and a Mali-G76 GPU). All executions used  8 threads on mobile CPUs, and  similarly 
all pipelines on mobile GPUs. 16-bit and 32-bit floating points are used for all GPU runs and CPU runs, respectively. 
All experiments were run 100 times but as the variance was very small, we only report  averages. 


\subsection{Overall Mobile Inference Evaluation}

Our comparison includes both  fusion rate\footnote{Fusion rate $=$ original layer count/fused layer count.} and execution latency.

\compactparagraph{Fusion rate.} Table~\ref{tab:eva_model_layer} shows detailed layer counts (including computation-intensive (CIL), memory-intensive (MIL), and all layers), and intermediate result sizes for models before fusion and after fusion with different frameworks. 
Note that \projectname is the only end-to-end framework that  can support all of the target models on both mobile CPU and mobile GPU. In  
Table~\ref{tab:eva_model_layer}, \PLDIDIFF{``-''} 
implies that this framework does not support this model. Certain 
extremely deep neural networks (e.g., Faster R-CNN and Masker R-CNN) are not supported by any other frameworks on mobile devices because these frameworks either 
lack  the support of multiple key operators and/or limited optimization supported in them lead to a  large model execution footprint.
For transformer-based models, only TFLite can support execution on mobile CPU (without GPU support).

Table~\ref{tab:eva_model_layer} shows that compared with the other frameworks, \projectname results in better fusion rates, with $1.3\times$ to $2.9\times$, $1.3\times$ to $8.1\times$, $1.3\times$ to $8.8\times$, and $1.3\times$ to $2.8\times$ over MNN, TVM, TFLite, and Pytorch, respectively.
Particularly, compared with original models, \projectname yields more benefits for R-CNN and Transformer-based models ($3.9\times$ to $10.0\times$ fusion rate) than 2D/3D CNNs ($1.7\times$ to $3.6\times$ fusion rate). 
This is because 2D/3D CNNs have higher fractions of computation-intensive layers that are in either One-to-Many or Many-to-Many types, while transformer-based models have more memory-intensive layers that are in One-to-One, Shuffle, or Reorganize categories.  The latter offers more fusion opportunities according to our mapping type analysis (Table~\ref{tab:mapping-type-lookup}).
Because of the same reason, 3D CNNs have the lowest fusion rate because they are more compute-intensive.
Moreover, comparing to TVM (that performs the best among all other frameworks), \projectname particularly yields more benefits for transformer-based models. This is because these models have more types of operators, and TVM's fixed pattern-based fusion cannot capture fusion opportunities among many types of operators while \projectname can. This result demonstrates that \projectname has a better generality.

\compactparagraph{Execution latency.} 
Table~\ref{tab:eva_performance_report} shows the execution latency evaluation results.
Comparing with MNN, TVM, TFLite, and Pytorch, with fusion optimization, \projectname achieves the speedup of   $1.4\times$ to $2.6\times$, $1.5\times$ to $3.5\times$, $1.4\times$ to $3.3\times$, and $1.6\times$ to $9.3\times$, respectively, 
on the mobile CPU. 
Focusing on mobile GPU, improvements  over  MNN, TVM, and TFLite are 
 $1.7\times$ to $2.6\times$, $2.0\times$ to $3.1\times$, $1.6\times$ to $4.0\times$, respectively, 
 whereas Pytorch does not support execution on this  mobile GPU.  
The reason for speedups including the fact that 
our baseline implementation  with a fixed-pattern fusion ({\tt OurB+}) is already faster than other frameworks (with speedup of $1.04\times$ to $5.2\times$ on mobile CPU and $1.05\times$ to $1.7\times$ on mobile GPU), and with our more advanced fusion, \projectname achieves $1.4\times$ to $2.5\times$ speedup over {\tt OurB+} on mobile CPU and $1.5\times$ to $3.9\times$ speedup on mobile GPU. 
In addition, comparing \projectname (with fusion) and our baseline without fusion ({\tt OurB}), our advanced fusion brings $1.5\times$ to $5.8\times$ speedup.
Moreover, fusion optimization offered by \projectname brings more benefits for mobile GPU than CPU, particularly for extremely deep models (e.g., Faster R-CNN and GPT-2). This is because mobile GPU offers more parallelism  but smaller cache capacity compared  to CPU,  and GPU kernel launch incurs certain overheads, so it is more sensitive to intermediate data reduction, kernel launch reduction, and processor utilization increase that are brought by \projectname's fusion.

\begin{figure}[t]
    \centering
    \includegraphics[width=1 \columnwidth]{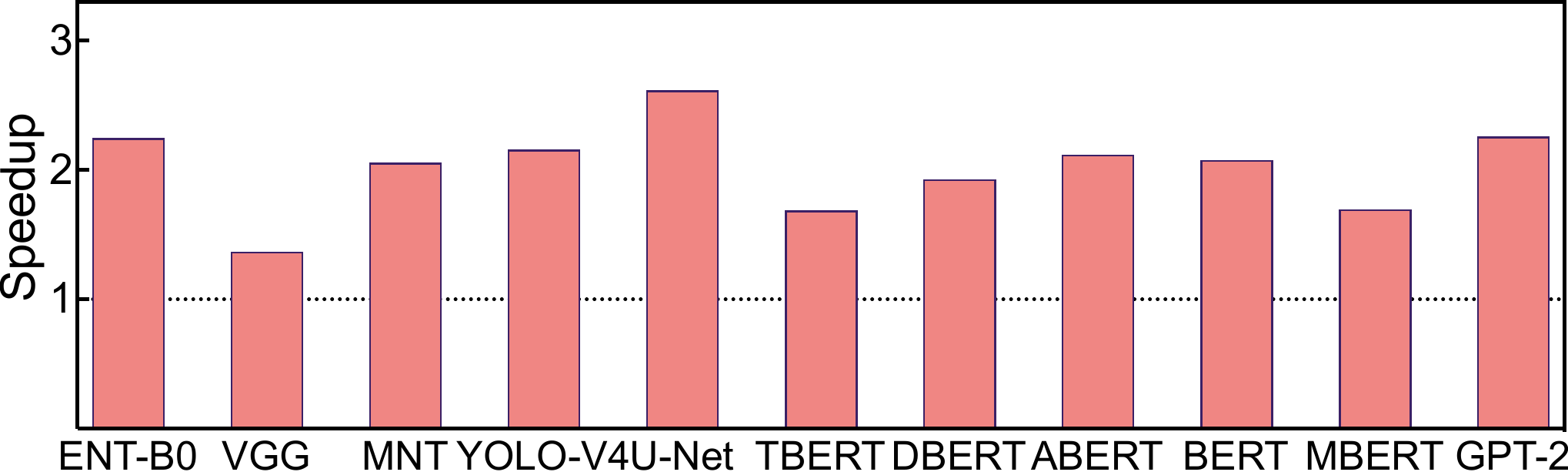}
    \caption{\PLDIDIFF{\textbf{Speedup over TASO optimized execution  on mobile CPU.} The models (computational graphs) are optimized by TASO and then  executed on TFLite.}}
    \label{fig:eva_taso}
\end{figure}

\PLDIDIFF{To further validate \projectname's performance, Figure~\ref{fig:eva_taso} compares it with a state-of-the-art computational graph substitution approach mentioned earlier, TASO~\cite{jia2019taso}. 
We  use TASO to optimize all eleven models (computational graphs) supported by TFLite among 
those listed in Table \ref{tab:eva_performance_report}, including EfficientNet-B0 (ENT-B0), VGG-16 (VGG), MobileNetV1-SSD (MNT), YOLO-V4, U-Net, TinyBERT (TBERT), DistilBERT (DBERT), ALBERT (ABERT), BERT$_{Base}$ (BERT), MobileBERT (MBT), and GPT-2. 
Then, for our experiments, these models are executed under  TFLite on mobile CPU (not GPU because TFLite lacks GPU 
support for many of these models).
Compared with TASO, \projectname yields $1.4\times$ to $2.6\times$ speedup on mobile CPU. 
The graph rewriting in \projectname is designed to work in conjunction with operator fusion and identifies a set of operators and rules for that specific purpose, thus enabling more fusion opportunities. TASO does not emphasize the relationship between graph rewriting and fusion,  resulting in less efficient execution as compared to   \projectname. }

\subsection{Understanding Fusion Optimizations}
This section studies the effect of our key optimizations. 

\begin{figure}[t]
    \centering
        \subfloat[CPU.]{
            \includegraphics[width=0.475\columnwidth]{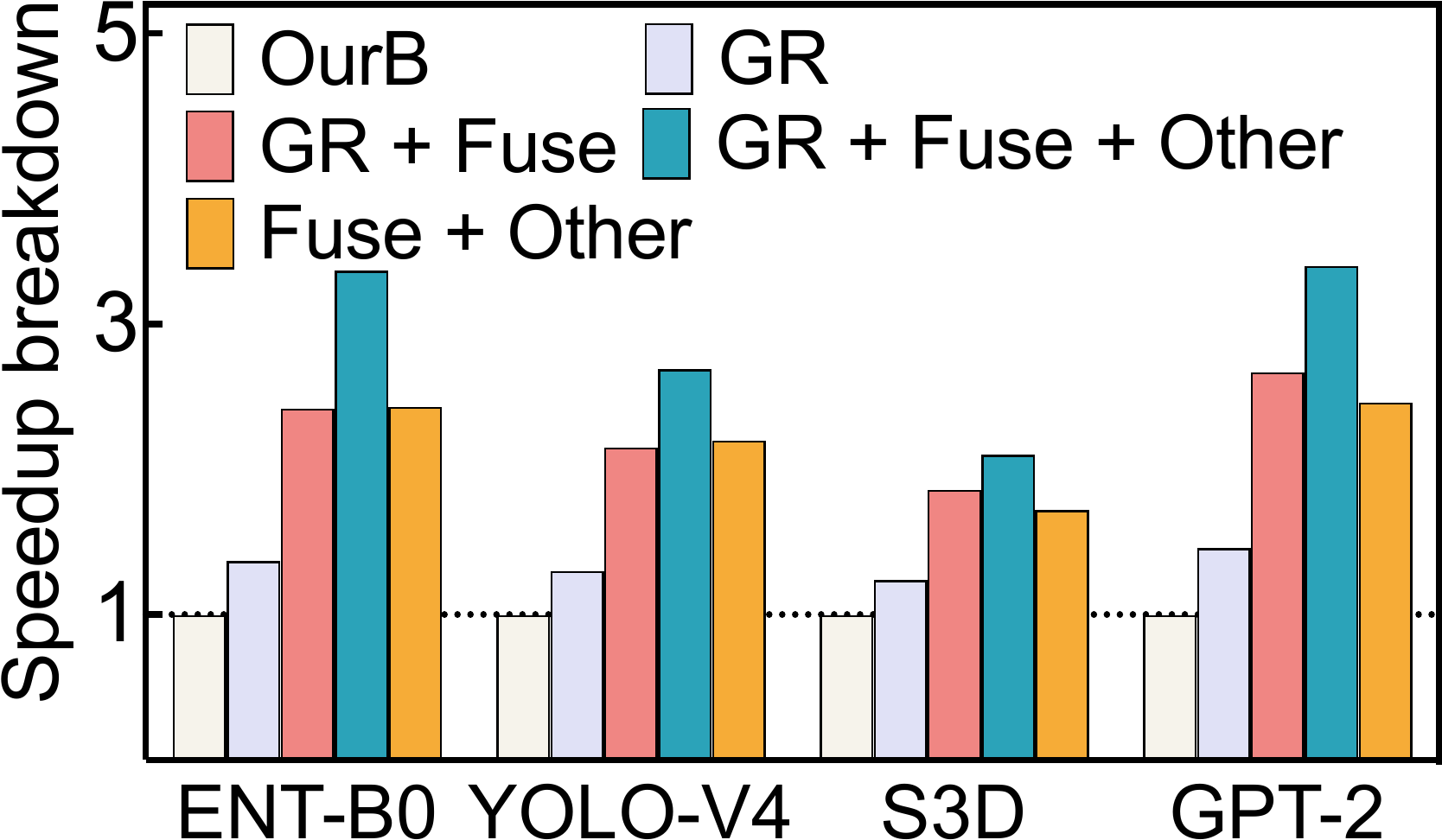}
        }
        \subfloat[GPU.]{
            \includegraphics[width=0.475\columnwidth]{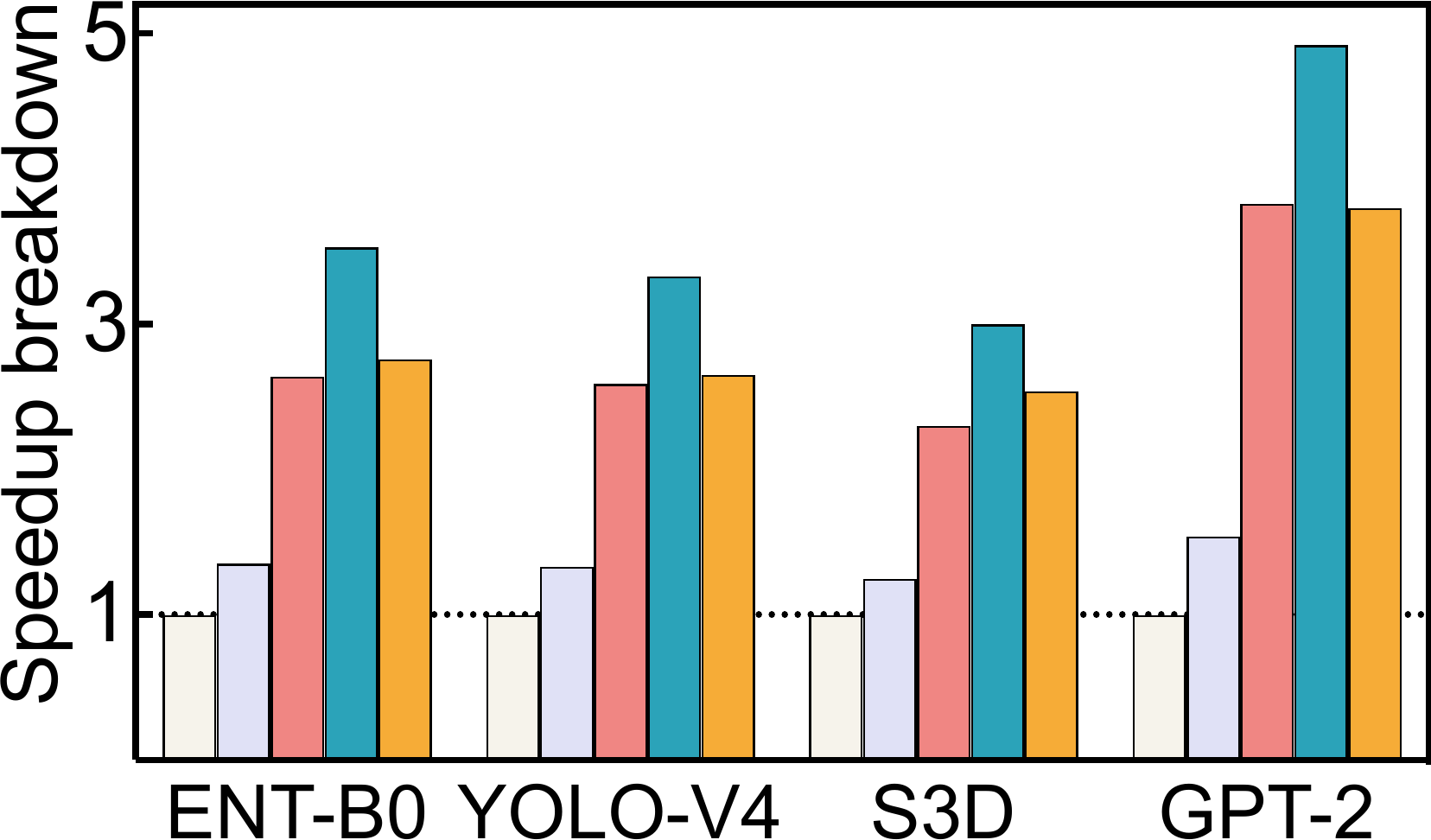}
        }
        \caption{\textbf{\PLDIDIFF{Optimization breakdown on y-axis}: speedup over OurB, i.e. a version w/o fusion opt.} GR, Fuse, and Other denote graph rewriting, fusion, and other fusion-related optimizations, respectively.}
    \label{fig:eva_opt_breakdown}
\end{figure}

\compactparagraph{Optimization breakdown.}
Figure~\ref{fig:eva_opt_breakdown} shows the impact of our proposed optimizations on latency with four models (EfficientNet-B0 (ENT-B0), YOLO-V4, S3D, and GPT-2 ) on both mobile CPU and GPU. Experiments on other models show a similar trend and are omitted due to space constraints.
We evaluate each compiler-based optimization speedup incrementally over the {\tt OurB} version.
Compared with {\tt OurB}, graph rewriting brings $1.2\times$ to $1.5\times$ speedup, fusion brings additional $1.6\times$ to $2.2\times$ speedup, and other optimizations \PLDIDIFF{(intra-block optimizations like data movement operator optimizations and inter-block optimizations like data format optimizations introduced in Section~\ref{sec:other-opt})} bring additional $1.3\times$ to $1.8\times$ speedup on mobile CPU. 
On mobile GPU, these numbers are $1.3\times$ to $1.5\times$, $2.1\times$ to $3.3\times$, and $1.7\times$ to $2.1\times$, respectively. 
Again, our fusion brings more benefit for mobile GPU than CPU due to the aforementioned reasons of memory, parallelism, and kernel launch overheads. 
Although graph rewriting by itself brings fewer benefits than the fusion, its hidden benefit is in enabling more fusion opportunities. Take GPT-2 as an example -- 
without graph rewriting, the generated fused layer count is 310, while after rewriting, it is 254 ($18\%$ reduction). This is because graph rewriting simplified the computational graph structure.
\PLDIDIFF{To further assess this impact of graph rewriting on operator fusion, the last bar (in orange) of Figure~\ref{fig:eva_opt_breakdown} reports the speedup of fusion with other optimizations only (i.e., without graph rewriting) over the baseline {\tt OurB}. Compared with no graph rewriting, graph rewriting brings an additional $1.4\times$ to $1.9\times$ and $1.6\times$ to $2.0\times$ speedup (over {\tt OurB}) on mobile CPU and GPU, respectively.}



\begin{figure}[t!]
    \centering
    \includegraphics[width=0.98 \columnwidth]{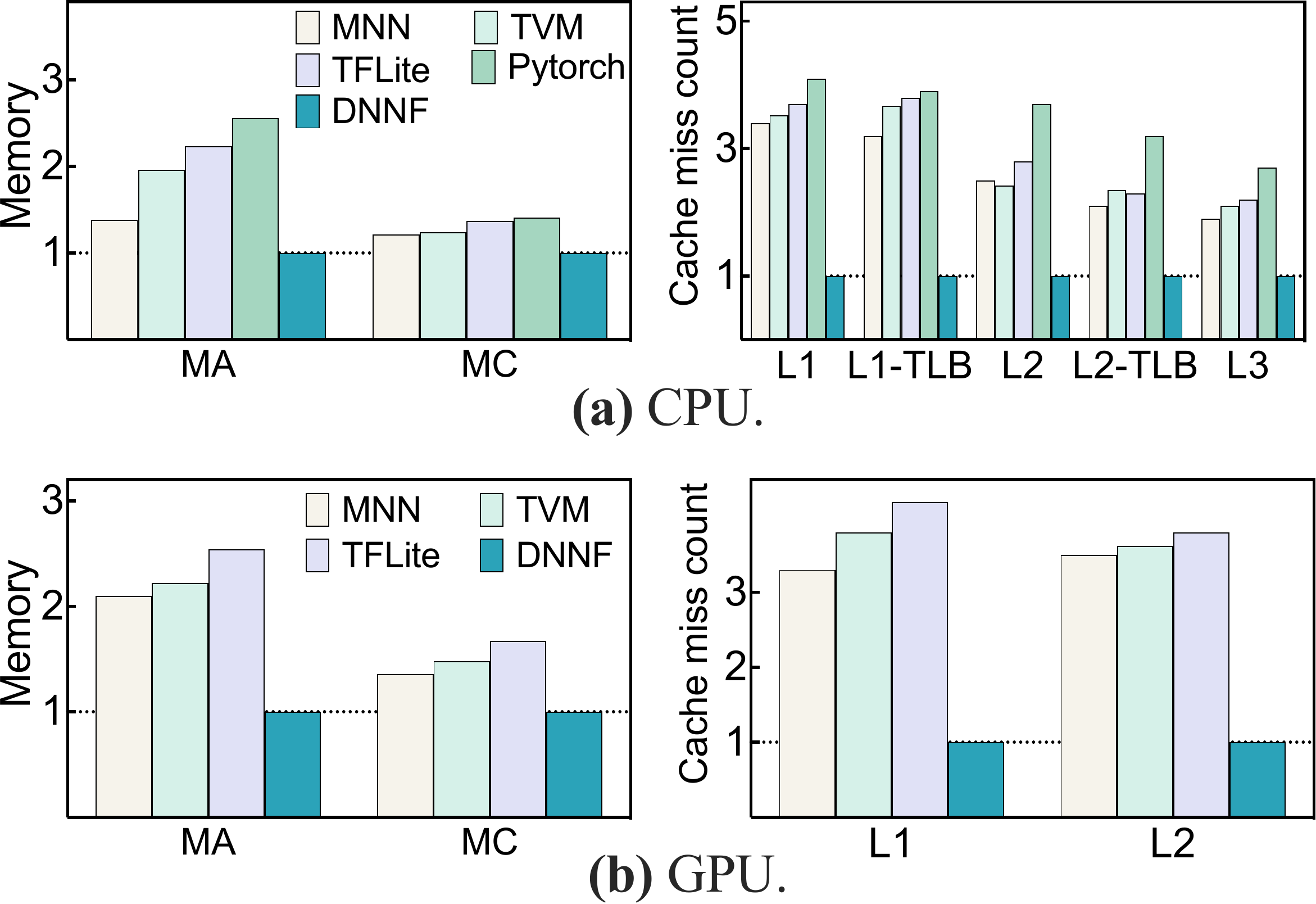}
    \caption{{\bf Memory (left) and cache miss (right) analysis.} MA and MC denote memory access and memory consumption, respectively. Cache miss count is compared on L1/L2/L3 data cache and L1/L2 TLB cache on mobile CPU, and  on L1/L2 data cache only on mobile GPU. All values are normalized w.r.t DNNF (the optimal version).}
    \label{fig:eva_cache_perf_all}
\end{figure}

\compactparagraph{Memory and cache performance}
We compare memory (and cache) performance of \projectname with other frameworks on both mobile CPU and GPU. We only report YOLO-V4's results due to space constraint, and because it is one of the models supported by all frameworks.
\PLDIDIFF{Figure~\ref{fig:eva_cache_perf_all} (a) left shows memory performance (measured in total memory accesses (MA) and memory consumption (MC)) -- MA and MC are collected from Snapdragon Profiler \cite{snapdragon-profiler}}, and Figure~\ref{fig:eva_cache_perf_all} (a) right shows cache miss count on data cache and TLB cache on mobile CPU. All values are normalized with respect to \projectname (i.e., our best  version) for readability. \projectname outperforms other frameworks on both memory access count and memory consumption because it eliminates materialization of more intermediate results.
Figure~\ref{fig:eva_cache_perf_all} (b) shows similar results on mobile GPU (excluding PyTorch because it does not support YOLO-V4 on mobile GPU). 
Mobile GPU results are generally better than CPU because mobile GPU has smaller cache capacity and simpler hierarchy (GPU only has L1/L2 v.s., CPU has L1/L2/L3). 
Thus, intermediate results reduction leads to more gains on mobile GPU.


\begin{figure}[t]
    \centering
        \subfloat[CPU and GPU utilization.]{
            \includegraphics[width=0.475\columnwidth]{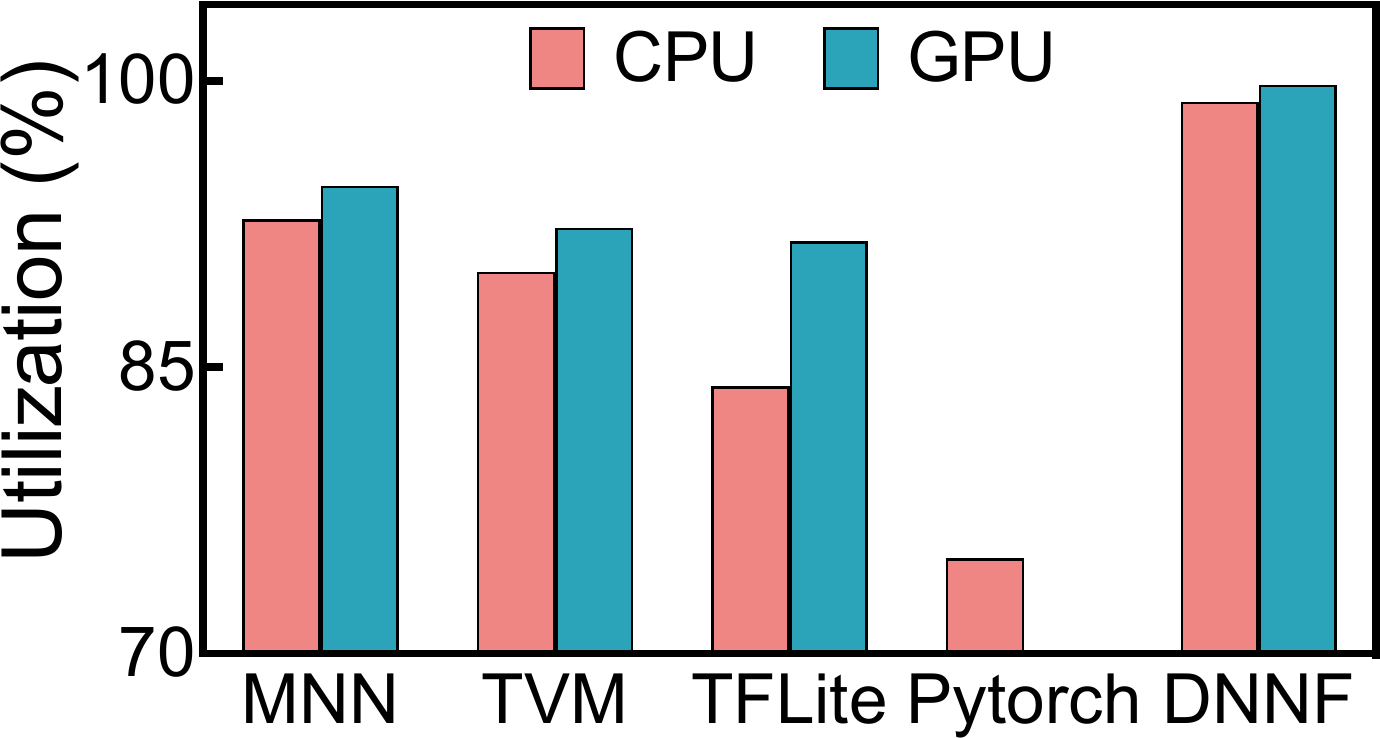}
        }
        \subfloat[Compilation time.]{
            \includegraphics[width=0.48\columnwidth]{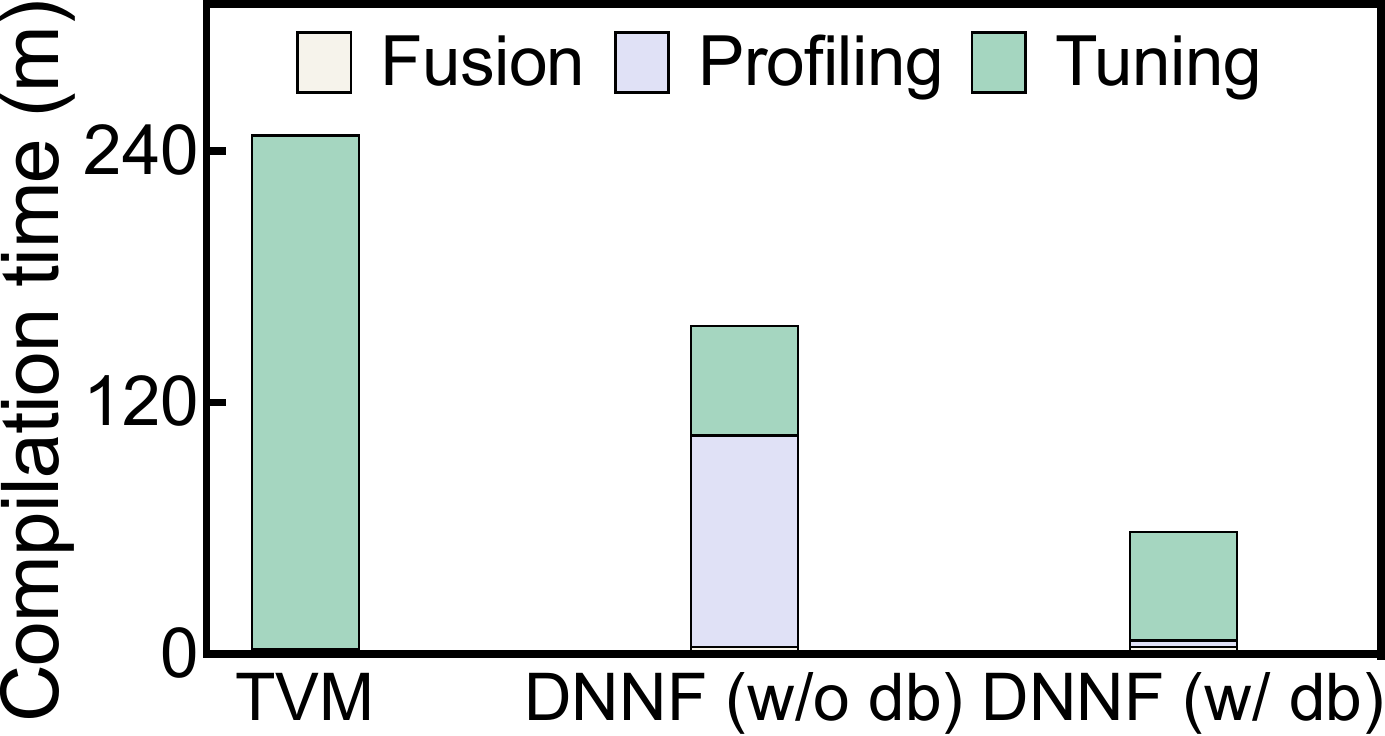}
        }
        \caption{{\bf (a) Mobile CPU and GPU utilization.} CPU utilization is averaged on 8-cores. {\bf (b) Compilation time.} Comparison between TVM and DNNF for YOLO-V4 on mobile CPU. DNNF (w/o db) is without the presence of an existing profiling database;  DNNF (w/ db) assumes such a database is pre-computed. \PLDIDIFF{{\tt Fusion} is invisible as it spends very little time on both TVM and DNNF.}}
    \label{fig:eva_utilization}
\end{figure}


\compactparagraph{CPU/GPU utilization.}
Figure~\ref{fig:eva_utilization} (a) reports the mobile CPU and GPU utilization on YOLO-V4. This result is collected by Snapdragon Profiler ~\cite{snapdragon-profiler}. It shows that \projectname results in the highest utilization on both CPU and GPU because its aggressive fusion groups more computation together, resulting in coarser-grained execution with less loop and parallel scheduling (and kernel launch for GPU only)  overhead. Particularly, the GPU utilization is slightly higher than CPU because of its higher parallelism.

\compactparagraph{Compilation time.} Figure~\ref{fig:eva_utilization} (b) compares the compilation time between TVM and \projectname for YOLO-V4 on mobile CPU. TVM's compilation time consists of operator fusion and other compiler optimizations ({\tt Fusion}), and tuning for performance-critical parameters, e.g., tiling size, unrolling factors, and others   ({\tt Tuning}). 
{\tt Tuning} dominates its compilation time, lasting for around 4 hours for YOLO-V4 on mobile CPU. \projectname's compilation time consists of operator fusion and other compiler optimizations ({\tt Fusion}), profiling to analyze the fusion benefits ({\tt Profiling}), and performance tuning ({\tt Tuning}). 
\PLDIDIFF{\projectname's {\tt Tuning} relies on an auto-tuner based on Genetic Algorithm (reported in 
our previous publication~\cite{niu2020patdnn})
to generate the exploration space. Compared with AutoTVM~\cite{chen2018learning}, our auto-tuning's  Genetic Algorithm allows  parameter search to start  with initializing an arbitrary number of chromosomes.}
Without our pre-existing 
profiling database, {\tt Profiling} and {\tt Tuning} dominate the compilation, requiring around 3 hours. With a pre-computed  database, {\tt Profiling} becomes very fast, and only {\tt Tuning} dominates the compilation, requiring around 1 hour. 
\PLDIDIFF{After evaluating 15 models in Table ~\ref{tab:eva_model_layer}, the profiling database consists of around 22K profiling entries with each one including operators' information (e.g., operator types, shape, and their combinations) and the latency 
achieved.}

\subsection{Portability}
\begin{figure}[t!]
    \centering
    \includegraphics[width=0.99 \columnwidth]{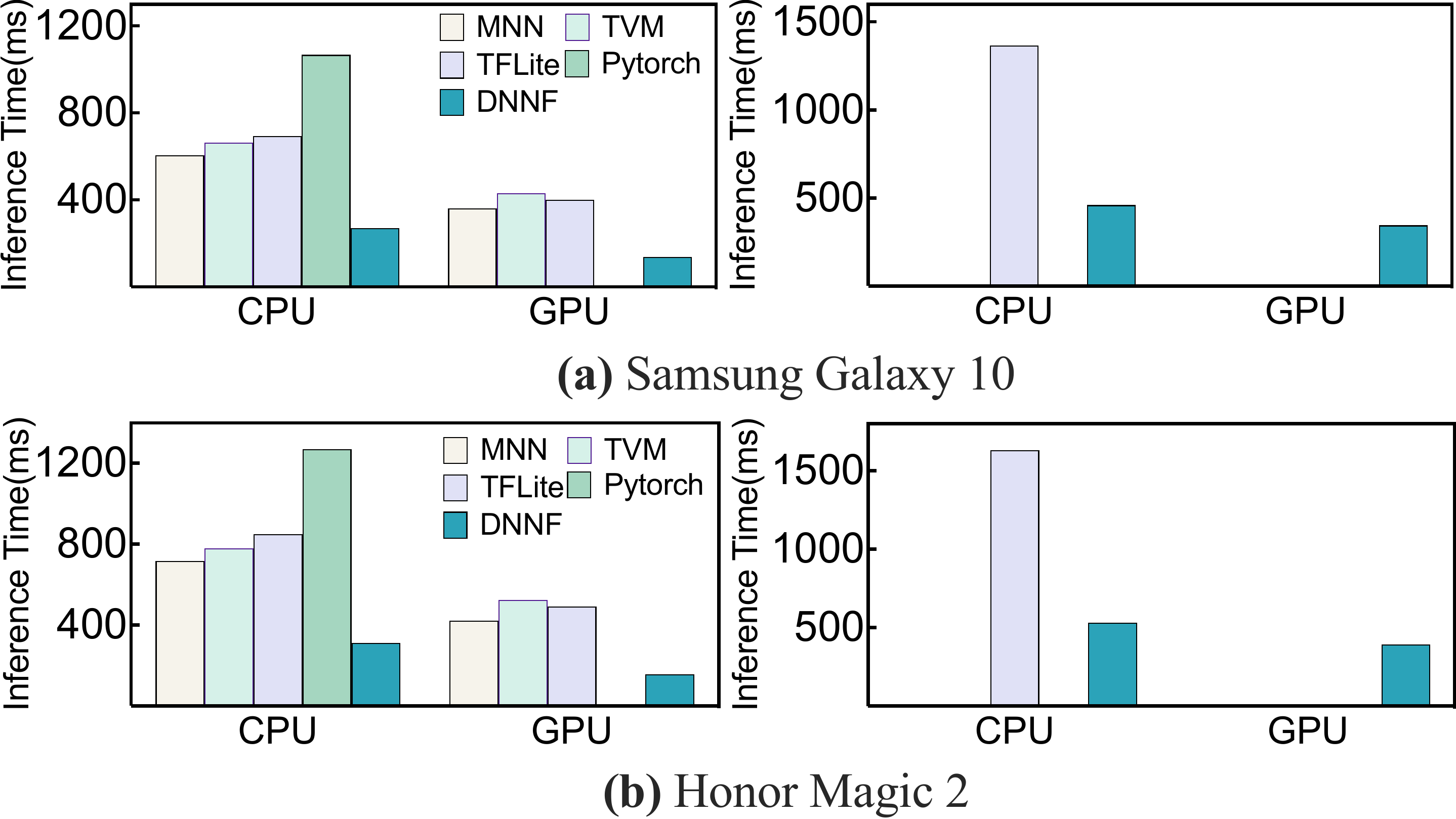}
    \caption{\textbf{Portability evaluation.} It is on Samsung Galaxy S10 and Honor Magic 2. Left two figures are YOLO-V4 and right two are GPT-2. Only TFLite supports GPT-2 on mobile CPU (no mobile GPU support).}
    \label{fig:eva_portibility}
\end{figure}

Figure~\ref{fig:eva_portibility} shows the execution latency on additional cell phones (Samsung Galaxy S10 and Honor Magic 2) to demonstrate effective portability. 
Only YOLO-V4 and GPT-2 are reported due to limited space. Other models show similar trends. 
In particular, \projectname shows a more stable performance on older generations of mobile devices. This is because our fusion significantly reduces the overall number of layers and intermediate result size, and older cell phones with more restricted resources are more sensitive to these.

\section{Related Work}


\compactparagraph{Operator fusion in end-to-end mobile DNN frameworks.} Operator fusion is an important optimization in many state-of-the-art end-to-end mobile DNN execution frameworks that are based on computational graphs, such as MNN~\cite{Ali-MNN}, TVM~\cite{chen2018tvm}, TensorFlow-Lite~\cite{TensorFlow-Lite}, and Pytorch~\cite{Pytorch-Mobile}. 
However, they all employ fixed-pattern fusion that is too restricted to cover diverse operators and layer connections in deep models like BERT --  \PLDIDIFF{for example, {\tt ConvTranspose} + {\tt ReLU} + {\tt Concat} cannot be recognized in TVM as 
it is not one of the specified patterns. Other examples that can be handled by \projectname and  cannot be recognized by TVM include {\tt MatMul} + {\tt Reshape} + {\tt Transpose} + {\tt Add} in GPT-2, and {\tt Sub} + {\tt Pow} + {\tt ReduceMean} + {\tt Add} + {\tt Sqrt} in TinyBERT}.
Comparing to these frameworks, \projectname works by classifying both the operators and their combinations, thus enabling a much larger set of optimizations.


\compactparagraph{Operator fusion on other ML frameworks.}
There are certain  recent frameworks that rely on polyhedral analysis to optimize DNN computations and support operator fusion.  
 R-Stream$\cdot$TF~\cite{pradelle2017polyhedral} shows a proof-of-concept adaptation of the R-Stream
polyhedral compiler to TensorFlow. 
Tensor Comprehensions~\cite{vasilache2018tensor} is an end-to-end compilation framework built on a domain-specific polyhedral analysis. These frameworks do not support mobile execution (i.e. 
ARM architecture), and thus we cannot  perform a direct  comparison between \projectname and them.  
As we have stated earlier, 
 \projectname maintains an operator view but builds a  higher-level abstraction on them.  In the future, we can  combine  \projectname's high-level abstraction to existing 
 domain-specific polyhedral analysis.  Similarly, another  promising direction will be  to integrate \projectname into other compilation-based DNN frameworks~\cite{gopinath2019compiling,mogers2020automatic} or other popular general tensor/matrix/linear algebra computation frameworks, such as MLIR~\cite{lattner2020mlir}, \PLDIDIFF{Tiramisu}~\cite{baghdadi2019tiramisu},
TACO~\cite{kjolstad2017tensor, kjolstad2019tensor}, Halide~\cite{halide}, and LGen~\cite{spampinato2014basic,kyrtatas2015basic}.

There also exist several other frameworks to optimize machine learning with operator fusion or fusion-based ideas. Closely related to \projectname -- 
Rammer~\cite{rammer-osdi20} relies on fix-pattern operator fusion to further reduce  kernel launch overhead of their optimized scheduling, 
Cortex~\cite{fegade2020cortex} proposes a set of optimizations based on kernel fusion for dynamic recursive models, 
\PLDIDIFF{TensorFlow XLA~\cite{tensorflow-xla} offers a more general fusion method than fix-pattern operator fusion by supporting reduce operations and element-wise operations, 
and TensorFlow Grapper ~\cite{tensorflow-grappler} provides an arithmetic optimizer that performs rewrites to achieve both fusion and arithmetic expression simplification (e.g., $a\times b+a\times c = a\times(b+c)$). Comparing with these frameworks, \projectname works by  classifying the operators and their combinations into several mapping categories, thus resulting in a more aggressive fusion plan and more performance gains. }
Elgamal~\cite{elgamal2017spoof} and Boehm~\cite{boehm2018optimizing} presently optimize general machine learning algorithms (e.g., SVM and  Kmeans) with operator fusion. 
These efforts have  both different targets and techniques compared to \projectname.

\compactparagraph{Polyhedral-based and other loop fusion methods.}  
Polyhedral analysis~\cite{pouchet2008iterative,bondhugula2008practical,yuki2012alphaz,kong2013polyhedral,doerfert2015polly,chatarasi2015polyhedral} is a prominent 
approach that offers a general and rigorous foundation for loop transformation and optimization. 

Many existing efforts~\cite{bondhugula2010model,acharya2018polyhedral,acharya2020effective} rely on a general polyhedral analysis to achieve optimized loop fusion. \PLDIDIFF{ Pouchet {\em et al.}~\cite{pouchet2011loop} have demonstrated  that polyhedral analysis can decompose the loop optimization problem into sub-problems  that 
have much lower complexity, enabling optimal selection. 
The problem arising because of a large number of operators in our target applications (models) is quite 
different, and thus there does not seem to be a direct application of Pouchet {\em et al.}'s  approach in our context.  } 
There have also been other  loop fusion efforts targeting   general programs~\cite{debray1988unfold,megiddo1997optimal,kennedy1993maximizing,kandemir1998improving}. 
In contrast to these general efforts, \projectname is more domain-specific, leveraging the knowledge of DNN computations with a higher-level abstraction to explore more aggressive loop fusion opportunities. 

\section{Conclusions and Future Work}\label{sec:conclusion}

This paper has presented  a new loop fusion framework called \projectname. The key advantages of \projectname include: 1) a new high-level abstraction comprising  mapping type of 
operators and their combinations and the  Extended Computational Graph, and analyses on these  abstractions, 2) a novel mathematical-property-based graph rewriting, and 3) an integrated fusion plan generation. \projectname is extensively evaluated on 15 diverse DNN models on multiple mobile devices, and evaluation results show that it outperforms four state-of-the-art DNN execution frameworks by up to $8.8\times$ speedup, and {\em \ for the first time} allows many cutting-edge DNN models not supported by prior end-to-end frameworks to execute on mobile devices efficiently (even in real-time).  
In addition, \projectname improves both cache performance and device utilization, enabling execution on devices with more restricted resources. It also reduces performance tuning time during  
compilation.

Our future work will enhance \projectname by combining  it with the latest model pruning advances~\cite{niu2020patdnn,DBLP:conf/dac/DongWNZLLGRLT20}. 
Though model pruning is effective,  with fusion the  dense versions are outperforming these efforts by having fewer layers. Thus, there is an opportunity to combine the two 
set of approaches to achieve an even better performance.

 
\begin{acks}
The authors would like to thank the anonymous reviewers for their valuable and thorough comments. The authors are especially grateful to the shepherd Tatiana Shpeisman for her innumerable helpful suggestions and comments that help improve this paper substantially. This work is supported in part by Jeffress Trust Awards in Interdisciplinary Research, and National Science Foundation (NSF) under CCF-1629392, CCF-2007793, CCF-1919117, and OAC-2034850. Any opinions, findings, and conclusions or recommendations expressed in this material are those of the authors and do not necessarily reflect the views of Thomas F. and Kate Miller Jeffress Memorial Trust, or NSF. 
\end{acks}




\end{document}